\documentclass[final]{cvpr}

\usepackage{times}
\usepackage{epsfig}
\usepackage{graphicx}
\usepackage{amsmath}
\usepackage{amssymb}
\usepackage{multirow}
\usepackage{booktabs}
\usepackage[noend]{algpseudocode}
\usepackage{algorithmicx,algorithm}
\usepackage{color}
\usepackage[dvipsnames]{xcolor}
\usepackage[numbers, sort]{natbib}
\usepackage{enumitem}
\usepackage[caption=false]{subfig}
\usepackage{makecell}
\usepackage{colortbl}
\usepackage[pagebackref=true,breaklinks=true,colorlinks,bookmarks=false,citecolor=ForestGreen]{hyperref}

\newcommand{\mAPK}{mAP$_\mathcal{K}$}
\newcommand{\APU}{AP$_\mathcal{U}$}

\newcommand{\gray}[1]{\textcolor{gray}{#1}}
\definecolor{llgray}{gray}{0.95}
\newcommand{\cg}{\cellcolor{llgray}}

\def\cvprPaperID{3605} 
\def\confYear{CVPR 2022}
\def\confName{CVPR}

\begin{document}

\title{Expanding Low-Density Latent Regions for Open-Set Object Detection}

\author{Jiaming Han$^{1,2*}$, Yuqiang Ren$^3$, Jian Ding$^{1,2}$, Xingjia Pan$^3$, Ke Yan$^{3\dag}$, Gui-Song Xia$^{1,2\dag}$\\
$^1$NERCMS, School of Computer Science, Wuhan University\\
$^2$State Key Lab. LIESMARS, Wuhan University\\
$^3$YouTu Lab, Tencent\\
{\tt\small \{hanjiaming, jian.ding, guisong.xia\}@whu.edu.cn}\\ 
{\tt\small \{condiren, kerwinyan\}@tencent.com, xjia.pan@gmail.com}\\
}

\maketitle

\newcommand\blfootnote[1]{%
\begingroup
\renewcommand\thefootnote{}\footnote{#1}%
\addtocounter{footnote}{-1}%
\endgroup
}

\blfootnote{$^*$ Work done during internship at Tencent YouTu Lab.}
\blfootnote{$^\dag$ Corresponding author.}

\begin{abstract}
Modern object detectors have achieved impressive progress under the close-set setup.
However, open-set object detection (OSOD) remains challenging since objects of unknown categories are often misclassified to existing known classes.
In this work, we propose to identify unknown objects by separating high/low-density regions in the latent space, based on the consensus that unknown objects are usually distributed in low-density latent regions.
As traditional threshold-based methods only maintain limited low-density regions, which cannot cover all unknown objects, we present a novel Open-set Detector (OpenDet) with expanded low-density regions.
To this aim, we equip OpenDet with two learners, Contrastive Feature Learner (CFL) and Unknown Probability Learner (UPL).
CFL performs instance-level contrastive learning to encourage compact features of known classes, leaving more low-density regions for unknown classes;
UPL optimizes unknown probability based on the uncertainty of predictions, which further divides more low-density regions around the cluster of known classes.
Thus, unknown objects in low-density regions can be easily identified with the learned unknown probability.
Extensive experiments demonstrate that our method can significantly improve the OSOD performance, \emph{e.g.}, OpenDet reduces the Absolute Open-Set Errors by 25\%-35\% on six OSOD benchmarks.
Code is available at: \url{https://github.com/csuhan/opendet2}.
\end{abstract}
\section{Introduction}
\label{sec:intro}
\begin{figure}[t]
    \centering
    \includegraphics[width=\linewidth]{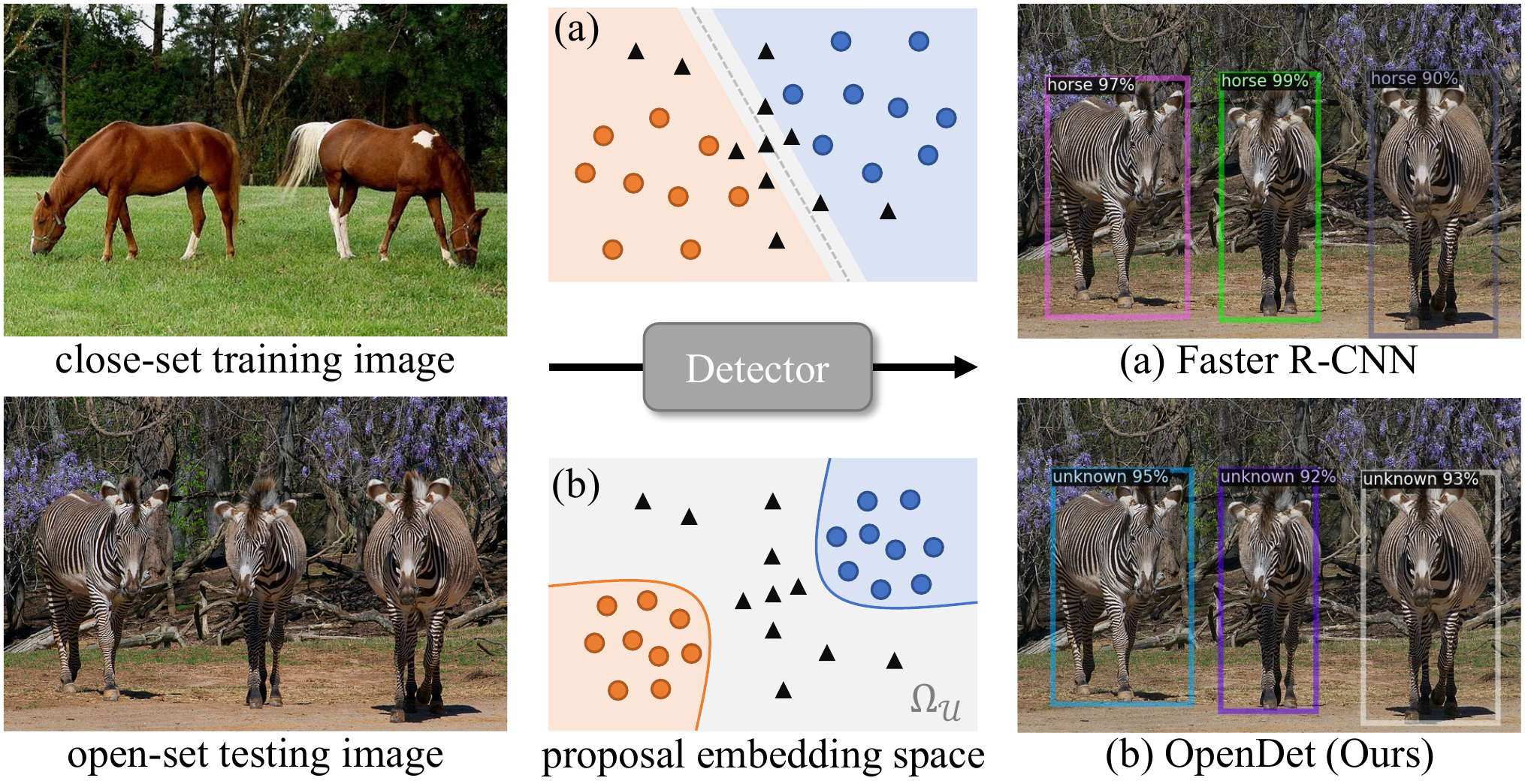}
    \caption{Trained on close-set images, 
    {\bf (a) threshold-based methods}, \emph{e.g.}, Faster R-CNN, usually misclassify unknown objects (black triangles, \emph{e.g.} zebra) into known classes (colored dots, \emph{e.g.} horse) due to limited low-density regions (in gray color). 
    {\bf (b) Our method} identifies unknown objects by expanding low-density regions. We encourage compact proposal features and learn clear separation between known and unknown classes.}
    \label{fig:mini_framework}
    \vspace{-3mm}
\end{figure}

Although the past decade has witnessed significant progress in object detection~\cite{girshick2014rich, ren2017faster, redmon2016you, lin2017focal, tian2019fcos, carion2020end}, modern object detectors are often developed with a close-set assumption that the object categories appearing in the testing process are contained by the training sets, 
and quickly lose their efficiency when handling real-world scenarios as many objects categories have never been seen in the training. 
See Fig.~\ref{fig:mini_framework} for an instance, where a representative object detector, \ie, Faster R-CNN~\cite{ren2017faster} trained on PASCAL VOC~\cite{everingham2010pascal}, misclassifies \texttt{zebra} into \texttt{horse} with high confidence, as the new class \texttt{zebra} is not contained by PASCAL VOC.
To alleviate this issue, Open-Set Object Detection (OSOD) has been recently investigated, where the detector trained on close-set datasets is asked to detect all known objects and identify unknown objects in open-set conditions. 

OSOD can be seen as an extension of Open-Set Recognition (OSR)~\cite{scheirer2012toward}.
Although OSR has been extensively studied~\cite{scheirer2012toward,bendale2016towards,ge2017generative,yoshihashi2019classification,chen2020learning,zhou2021learning}, rare works attempted to solve the challenging OSOD.
Dhamija~\emph{et al.}~\cite{dhamija2020overlooked} first benchmarked the open-set performance of some representative methods~\cite{ren2017faster,redmon2016you,lin2017focal}, which indicates most detectors are overestimated in open-set conditions.
Miller~\emph{et al.}~\cite{miller2018dropout, miller2019evaluating} adopt dropout sampling~\cite{gal2016dropout} to improve the robustness of detectors in open-set conditions.
Joseph~\emph{et al.}~\cite{joseph2021towards} proposed an energy-based unknown identifier by fitting the energy distributions of known and unknown classes.
In summary, prior works usually leverage hidden evidence (\emph{e.g.}, the output logits) of pre-trained models as unknown indicators, with the cost of additional training step and complex post-processing.
Can we train an open-set detector \emph{with only} close-set data, and \emph{directly} apply it to open-set environments \emph{without} complex post-processing?

We draw inspiration from the consensus that known objects are usually clustered to form high-density regions in the latent space, while unknown objects (or novel patterns) are distributed in low-density regions~\cite{grandvalet2004semi,chapelle2006semi,ren2018meta}.
From this perspective, proper separation of high/low-density latent regions is crucial for unknown identification.
However, traditional methods, \emph{e.g.}, hard-thresholding (Fig.~\ref{fig:mini_framework} (a)), only maintain limited low-density regions, as higher thresholds will hinder the close-set accuracy.
In this work, we propose to identify unknown objects by expanding low-density latent regions (Fig.~\ref{fig:mini_framework} (b)).
Firstly, we learn compact features of known classes, leaving more low-density regions for unknown classes.
Then, we learn an unknown probability for each instance, which serves as a threshold to divide more low-density regions around the cluster of known classes.
Finally, unknown objects distributed in these regions can be easily identified.

More specifically, we propose an Open-set Detector (OpenDet) with two learners, Contrastive Feature Learner (CFL) and Unknown Probability Learner (UPL), which expands low-density regions from two folds.
Let us denote the latent space with $\Omega=\Omega_\mathcal{K} \cup \Omega_\mathcal{U}$, where $\Omega_\mathcal{K}$ and $\Omega_\mathcal{U}$ represent high/low-density sub-space, respectively.
CFL performs instance-level contrastive learning to encourage intra-class compactness and inter-class separation of known classes, which expands $\Omega_\mathcal{U}$ by narrowing $\Omega_\mathcal{K}$.
UPL learns unknown probability for each instance based on the uncertainty of predictions.
As we carefully optimize UPL to maintain the close-set accuracy, the learned unknown probability can serve as a threshold to divide more $\Omega_\mathcal{U}$ around $\Omega_\mathcal{K}$.
In the testing phase, we directly classify an instance into the \texttt{unknown} class if its unknown probability is the largest among all classes.

\begin{figure}[!t]
    \centering
    \includegraphics[width=\linewidth]{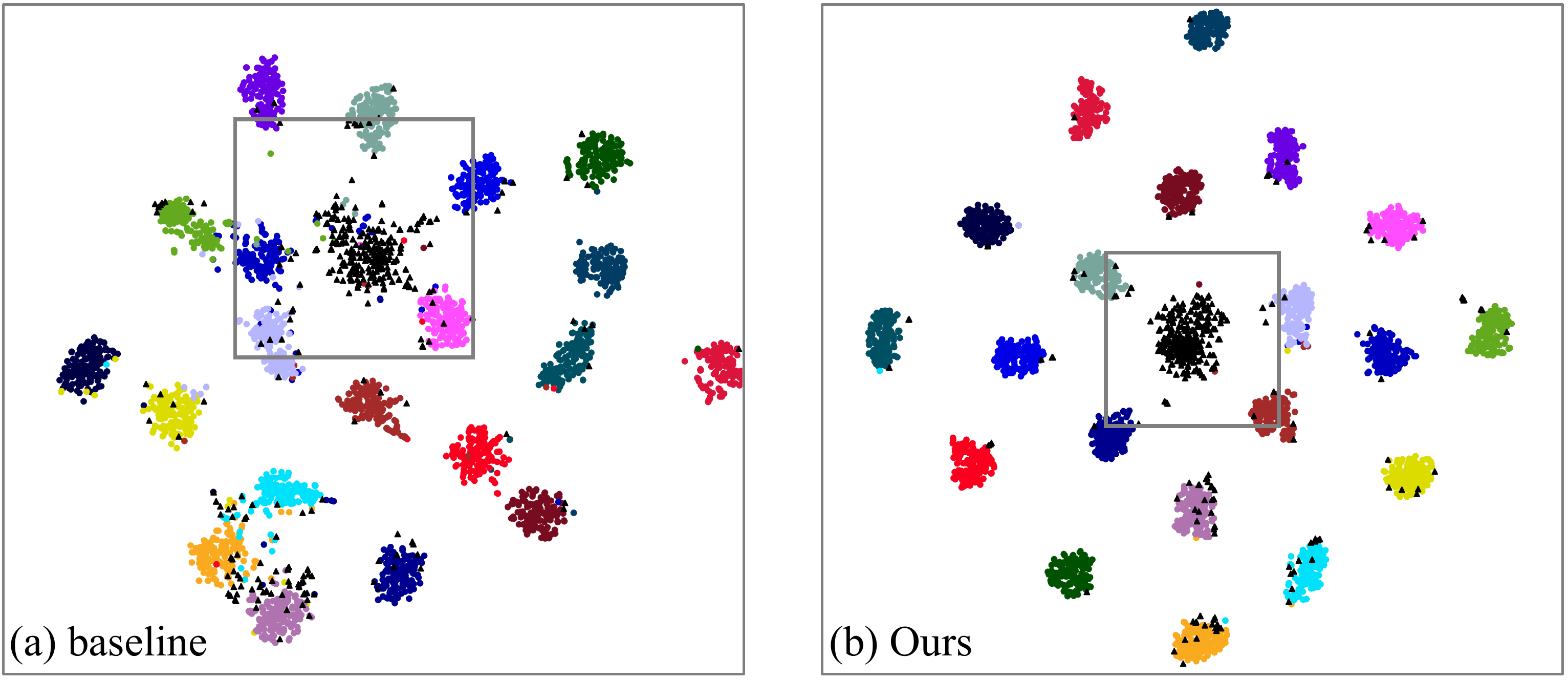}
    \caption{{\bf t-SNE visualization of latent features.} We take VOC classes as known classes (colored dots), and non-VOC classes in COCO as unknown classes (black triangles). Our method learns a clear separation between known and unknown classes.}
    \label{fig:tsne_vis}
    \vspace{-3mm}
\end{figure}

To demonstrate the effectiveness of our method, we take PASCAL VOC~\cite{everingham2010pascal} for close-set training and construct several open-set settings considering both VOC and COCO~\cite{lin2014microsoft}.
Compared with previous methods, OpenDet shows significant improvements on all open-set metrics without compromising the close-set accuracy.
For example, OpenDet reduces the Absolute Open-Set Errors (introduced in Sec.~\ref{subsec:exp_setup}) by 25\%-35\% on six open-set settings.
We also visualize the latent feature in Fig.~\ref{fig:tsne_vis}, where OpenDet learns clear separation between known and unknown classes.
Besides, we conduct extensive ablation experiments to analyze the effect of our main components and core design choices.
Furthermore, we show that OpenDet can be easily extended to one-stage detectors and achieve satisfactory results.
We summarize our contributions as:
\begin{itemize}[leftmargin=*]
    \vspace{-2mm}
    \item To our best knowledge, we are the first to solve the challenging OSOD by modeling low-density latent regions.
    \vspace{-2mm}
    \item We present a novel Open-set Detector (OpenDet) with two well-designed learners, CFL and UPL, which can be trained in an end-to-end manner and directly applied to open-set environments.
    \vspace{-2mm}
    \item We introduce a new OSOD benchmark. Compared with previous methods, OpenDet shows significant improvements on all open-set metrics, \emph{e.g.}, OpenDet reduces the Absolute Open-Set Errors by 25\%-35\%.
\end{itemize}
\section{Related Work}
\label{sec:related}

\begin{figure*}[!t]
    \centering
    \includegraphics[width=0.98\textwidth]{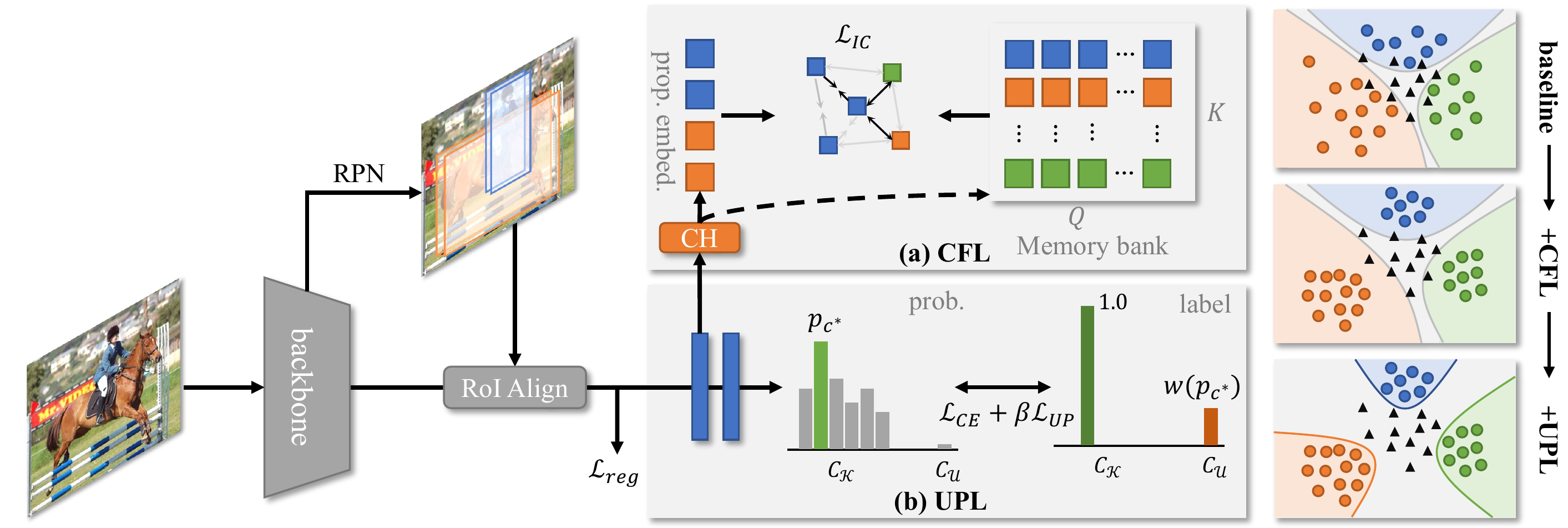}
    \caption{{\bf Overview of our proposed method.} 
    {\bf Left:} OpenDet is a two-stage detector with {\bf (a)} Contrastive Feature Learner (CFL) and {\bf (b)} Unknown Probability Learner (UPL). 
    {\bf CFL} first encodes proposal features into low-dimensional embeddings with the Contrastive Head (CH). Then we optimize these embeddings between the mini-batch and memory bank with an Instance Contrastive Loss $\mathcal{L}_{IC}$. 
    {\bf UPL} learns probabilities for both known classes $C_\mathcal{K}$ and unknown class $C_\mathcal{U}$ with cross-entropy loss $\mathcal{L}_{CE}$ and Unknown Probability Loss $\mathcal{L}_{UP}$. 
    {\bf Right:} A toy illustration of how different components work. Colored dots and triangles denote proposal features of different known and unknown classes, respectively. Our method identifies unknown by expanding low-density latent regions (in gray color).}
    \label{fig:main_framework}
    \vspace{-2mm}
\end{figure*}

\noindent {\bf Open-Set Recognition.}
Early attempts on OSR~\cite{scheirer2014probability,jain2014multi,zhang2016sparse,bendale2015towards,junior2017nearest} usually leverage traditional machine learning methods, \emph{e.g.}, SVM~\cite{scheirer2014probability, jain2014multi}.
Bendale~\emph{et al.}~\cite{bendale2016towards} introduced OpenMax, the first deep learning-based OSR method, which redistributes the output probabilities of the softmax layer.
Other approaches include generative adversarial network-based methods~\cite{ge2017generative,neal2018open} which generate potential open-set images to train an open-set classifier,
reconstruction-based methods~\cite{yoshihashi2019classification,oza2019c2ae,sun2020conditional} which adopt auto-encoder to recover latent features and identify unknown by reconstruction errors,
and prototype-based methods~\cite{chen2020learning,chen2021adversarial} which identify open-set images by measuring the distance to learned prototypes.
In addition, Zhou \emph{et al.}~\cite{zhou2021learning} proposed to learn data placeholders to anticipate open-set data and classifier placeholders to distinguish known and unknown.
Kong \emph{et al.}~\cite{kong2021opengan} utilized an adversarially trained discriminator to detect unknown examples.
Our method is more related to~\cite{zhou2021learning}. Differently, \cite{zhou2021learning} requires close-set pre-train and calibration on validation sets, while our method is trained in an end-to-end manner, and the learned unknown probability is accurate and calibration-free.

\noindent {\bf Open-Set Object Detection} is an extension of OSR in object detection.
Dhamija~\emph{et al.}~\cite{dhamija2020overlooked} first formalized OSOD and benchmarked some representative detectors by their classifiers. Classifiers with a background class~\cite{ren2017faster} performs better than one-vs-rest~\cite{lin2017focal} and objectness-based~\cite{redmon2016you} classifiers in handling unknown objects. 
Dhamija~\emph{et al.}~\cite{dhamija2020overlooked} also show that the performance of most detectors is overestimated in open-set conditions.
Miller~\emph{et al.}~\cite{miller2018dropout,miller2019evaluating} utilized dropout sampling~\cite{gal2016dropout} to estimate uncertainty in object detection and thus reduce open-set errors.
Joseph~\emph{et al.}~\cite{joseph2021towards} proposed an energy-based unknown identifier by fitting the energy distributions of known and unknown classes.
However, the approach in~\cite{joseph2021towards} requires extra open-set data of unknown classes, which violates the original definition of OSOD.
In summary, previous methods leverage hidden evidence (\emph{e.g.}, the output logits) of pre-trained models as unknown indicators. But they need additional training step and complex post-processing to estimate the unknown indicator.
In contrast, OpenDet can be trained with only close-set data and directly identify unknown objects with the learned unknown probability.

\noindent {\bf Contrastive Learning} is a methodology to learn representation by pulling together positive sample pairs while pushing apart negative sample pairs, which has been recently popularized for self-supervised representation learning~\cite{he2020momentum,chen2020simple,grill2020bootstrap,chen2021exploring,caron2020unsupervised,ding2021unsupervised}.
Khosla~\emph{et al.}~\cite{khosla2020supervised} first extended self-supervised contrastive learning to the full-supervised setting and received a lot of attention from other fields, \emph{e.g.}, long-tailed recognition~\cite{wang2021contrastive,cui2021parametric}, semantic segmentation~\cite{van2021unsupervised,wang2021exploring} and few-shot object detection~\cite{sun2021fsce}.
Our approach is also inspired by supervised contrastive learning~\cite{khosla2020supervised}.
In this work, we explore \emph{instance-level} contrastive learning to learn compact features of object proposals.

\noindent {\bf Uncertainty Estimation.} Neural networks tend to produce over-confident predictions~\cite{lakshminarayanan2016simple}. Estimating the uncertainty of model predictions is important for real-world applications. 
Currently, uncertainty estimation can be categorized into sampling-based and sampling-free methods. 
Sampling-based methods ensemble predictions of multiple runs~\cite{gal2016dropout} or multiple models~\cite{lakshminarayanan2016simple}, which are not applicable for speed-critical object detection.
Sampling-free methods learn additional confidence value~\cite{devries2018learning, sensoy2018evidential} to estimate uncertainty.
Our method belongs to the latter family. The learned unknown probability can reflects the uncertainty of predictions.
\section{Methodology}
\label{sec:method}

\subsection{Preliminary}
\label{subsec:preliminaries}
We formalize OSOD based on prior works~\cite{dhamija2020overlooked,joseph2021towards}.
Let us denote with $D=\{(x, y), x\in X, y \in Y\}$ an object detection dataset, where $x$ is an input image and $y=\{(c_i, \mathbf{b}_i)\}_{i=1}^N$ denotes a set of objects with corresponding class label $c$ and bounding box $\mathbf{b}$.
We train the detector on the training set $D_{tr}$ with $K$ known classes $C_\mathcal{K}=\{1, \dots, K\}$, and test it  on the testing set $D_{te}$ with objects from both known classes $C_\mathcal{K}$ and unknown classes $C_\mathcal{U}$.
The goal is to detect all known objects (objects $\in C_\mathcal{K}$), and identify unknown objects (objects $\in C_\mathcal{U}$) so that they will not be misclassified to $C_\mathcal{K}$.
As it is impossible to list infinite unknown classes, we denote them with $C_\mathcal{U}=K+1$.

Different from OSR, OSOD has its unique challenges.
In OSR, an image only belongs to $C_\mathcal{K}$ or $C_\mathcal{U}$; any example out of $C_\mathcal{K}$ is defined as unknown.
In OSOD, an image may contain objects from both $C_\mathcal{K}$ and $C_\mathcal{U}$, which is defined as mixed unknown~\cite{dhamija2020overlooked}.
That means unknown objects will also appear in $D_{tr}$ but have not been labeled yet.
Besides, detectors usually keep a background class $C_{bg}$ which is easily confused with $C_\mathcal{U}$.
\vspace{-0.5mm}

\subsection{Baseline Setup}
\label{subsec:baseline}
We setup the baseline with Faster R-CNN~\cite{ren2017faster}, which consists of a backbone, Region Proposal Network (RPN) and R-CNN. 
The standard R-CNN includes a shared fully connected (FC) layer and two separate FC layers for classification and regression.
We augment R-CNN in three ways. 
{\bf (a)} We replace the shared FC layer with two parallel FC layers so that the module applied to the classification branch will not affect the regression task.
{\bf (b)} Inspired by~\cite{chen2020learning,wang2020frustratingly}, we use cosine similarity-based classifier to alleviate the over-confidence issue~\cite{bendale2016towards,padhy2020revisiting}.
Specifically, we adopt scaled cosine similarity scores as output logits: $s_{i,j} = \frac{\alpha \mathcal{F}(x)_i^\top w_j}{\|\mathcal{F}(x)_i\| \|w_j\|}$, 
where $s_{i,j}$ denotes the similar score between $i$-th proposal features $\mathcal{F}(x)_i$ and weight vector of class $j$. $\alpha$ is the scaling factor ($\alpha$=20 by default).
{\bf (c)} The box regressor is set to class-agnostic, \emph{i.e.}, the regression branch outputs a vector of length 4 rather than $4(K+2)$.
Note that our baseline does not improve the open-set performance, but it is effective for the whole framework (Fig.~\ref{fig:main_framework}).

\subsection{Contrastive Feature Learner}
\label{subsec:cfl}
This section presents Contrastive Feature Learner (CFL) to encourage intra-class compactness and inter-class separation, which expands low-density latent regions by narrowing the cluster of known classes.
As shown in Fig.~\ref{fig:main_framework} (a), CFL contains a contrastive head (CH), a memory bank, and an instance contrastive loss $\mathcal{L}_{IC}$.
For a proposal feature $\mathcal{F}(x)_i$, we first encode it into a low-dimensional embedding with CH. Then, we optimize the embeddings from the mini-batch and memory bank with $\mathcal{L}_{IC}$.
We give more details in the following part.

\noindent {\bf Contrastive Head.} We build a contrastive head (CH) to map high-dimensional proposal feature $\mathcal{F}(x)_i$ to low-dimensional proposal embedding $\mathbf{z}_i \in \mathbf{R}^{d}$ ($d=128$ by default).
In detail, CH is a multilayer perceptron with sequential FC, ReLU, FC, and L2-Norm layers, which is applied to the classification branch of R-CNN in training and abandoned during inference.

\noindent {\bf Class-Balanced Memory Bank.} Popular contrastive representation learning usually adopts large-size mini-batch~\cite{khosla2020supervised} or memory bank~\cite{he2020momentum} to increase the diversity of exemplars.
Here we build a novel class-balanced memory bank to increase the diversity of object proposals. 
Specifically, for each class $c \in C_\mathcal{K}$, we initialize a memory bank $M(c)$ of size $Q$.
Then, we sample representative proposals from a mini-batch with two steps:
{\bf (a)} We sample proposals with Intersection of Union (IoU) $ > T_m$ where $T_m$ is an IoU threshold to ensure the proposals contain relevant semantics.
{\bf (b)} For each mini-batch, we sample $q (q \leq Q)$ proposals that are least similar (\emph{i.e.}, minimum cosine similarity) with existing exemplars in $M(c)$. This step makes our memory banks store more diverse exemplars and enable long-term memory.
Finally, we repeat (a) and (b) every iteration where the oldest proposals are out of the memory and the newest into the queue.

\noindent {\bf Instance-Level Contrastive Learning.} Inspired by supervised contrastive loss~\cite{khosla2020supervised}, we propose an Instance Contrastive (IC) Loss to learn more compact features of object proposals.
Assume we have a mini-batch of $N$ proposals, IC Loss is formulated as:
\begin{equation}\label{eq:ic_loss}
\mathcal{L}_{IC} = \frac{1}{N} \sum_{i=1}^N \mathcal{L}_{IC}(\mathbf{z}_i),
\end{equation}
\begin{equation}\label{eq:ic_loss_item}
\resizebox{.9\hsize}{!}{
    $\mathcal{L}_{IC}(\mathbf{z}_i)=\frac{1}{|M(\mathbf{c}_i)|} \sum\limits_{\mathbf{z}_{j} \in M(\mathbf{c}_i)} \log \frac{\exp \left(\mathbf{z}_{i} \cdot \mathbf{z}_{j} / \tau\right)}{\sum_{\mathbf{z}_{k} \in A(c_i)} \exp \left(\mathbf{z}_{i} \cdot \mathbf{z}_{k} / \tau\right)},$
}
\end{equation}
where $\mathbf{c}_i$ is the class label of $i$-the proposal, $\tau$ is a temperature hyper-parameter, $M(\mathbf{c}_i)$ denotes the memory bank of class $\mathbf{c}_i$, and $A(c_i)=M \backslash M(c_i)$.
Note that we only optimize proposals with IoU $ > T_b$ where $T_b$ is an IoU threshold similar to $T_m$.

Although unknown objects are unavailable in training, the separation of known classes benefits unknown identification.
Optimizing $\mathcal{L}_{IC}$ is equivalent to pushing the cluster of known classes away from low-density latent regions.
As shown in Fig.~\ref{fig:tsne_vis} (b), our method learns a clear separation between known and unknown classes with only close-set training data.

\begin{table*}[t]
    \centering
\small
\resizebox{\textwidth}{!}{%
\begin{tabular}{@{}l|c|cccc|cccc|cccc@{}}
    \midrule
    \multirow{2}{*}{Method} & VOC & \multicolumn{4}{c|}{VOC-COCO-20} & \multicolumn{4}{c|}{VOC-COCO-40} & \multicolumn{4}{c}{VOC-COCO-60} \\ \cmidrule(lr){2-2} \cmidrule(lr){3-6}  \cmidrule(lr){7-10} \cmidrule(lr){11-14}
                & mAP$_\mathcal{K \uparrow}$ & WI$_\downarrow$   & AOSE$_\downarrow$  & mAP$_\mathcal{K \uparrow}$   & AP$_\mathcal{U \uparrow}$  & WI$_\downarrow$   & AOSE$_\downarrow$  & mAP$_\mathcal{K \uparrow}$   & AP$_\mathcal{U \uparrow}$  & WI$_\downarrow$   & AOSE$_\downarrow$  & mAP$_\mathcal{K \uparrow}$   & AP$_\mathcal{U \uparrow}$  \\ \midrule
    FR-CNN~\cite{ren2017faster}    & 80.10 & 18.39 & 15118 & 58.45 & 0      & 22.74 & 23391  & 55.26 & 0     & 18.49 & 25472 & 55.83 & 0       \\ 
    FR-CNN$^*$~\cite{ren2017faster}& 80.01 & 18.83 & 11941 & 57.91 & 0      & 23.24 & 18257  & 54.77 & 0     & 18.72 & 19566 & 55.34 & 0       \\ 
    PROSER~\cite{zhou2021learning} & 79.68 & 19.16 & 13035 & 57.66 & 10.92  & 24.15 & 19831  & 54.66 & 7.62  & 19.64 & 21322 & 55.20 & 3.25    \\
    ORE~\cite{joseph2021towards}   & 79.80 & 18.18 & 12811 & 58.25 & 2.60   & 22.40 & 19752  & 55.30 & 1.70  & 18.35 & 21415 & 55.47 & 0.53    \\ 
    DS~\cite{miller2018dropout}    & 80.04 & 16.98 & 12868 & 58.35 & 5.13   & 20.86 & 19775  & 55.31 & 3.39  & 17.22 & 21921 & 55.77 & 1.25    \\ 
    \midrule
    OpenDet      & 80.02 & \bf{14.95} & \bf{11286} & \bf{58.75} & \bf{14.93} & \bf{18.23} & \bf{16800}  & \bf{55.83} & \bf{10.58} & \bf{14.24} & \bf{18250} & \bf{56.37} & \bf{4.36} \\ 
    \midrule
\end{tabular}%
}

    \caption{{\bf Comparisons with other methods on VOC and VOC-COCO-T$_1$.}
    We report close-set performance (\mAPK) on VOC, and both close-set (\mAPK) and open-set (WI, AOSE, \APU) performance of different methods on VOC-COCO-\{20, 40, 60\}.
    $^*$ means a higher score threshold (\emph{i.e.} 0.1) for testing.}
    \vspace{-2mm}
    \label{tab:voc_coco_a}
\end{table*}

\begin{table*}[t]
    \centering
\small
\resizebox{\textwidth}{!}{%
\begin{tabular}{@{}l|cccc|cccc|cccc@{}}
\midrule
\multirow{2}{*}{Method} & \multicolumn{4}{c|}{VOC-COCO-0.5$n$} & \multicolumn{4}{c|}{VOC-COCO-$n$} & \multicolumn{4}{c}{VOC-COCO-4$n$} \\ \cmidrule(lr){2-5}  \cmidrule(lr){6-9} \cmidrule(lr){10-13}
               & WI$_\downarrow$   & AOSE$_\downarrow$  & mAP$_\mathcal{K \uparrow}$   & AP$_\mathcal{U \uparrow}$  & WI$_\downarrow$   & AOSE$_\downarrow$  & mAP$_\mathcal{K \uparrow}$   & AP$_\mathcal{U \uparrow}$  & WI$_\downarrow$   & AOSE$_\downarrow$  & mAP$_\mathcal{K \uparrow}$   & AP$_\mathcal{U \uparrow}$  \\ \midrule
FR-CNN~\cite{ren2017faster}    & 9.25  & 6015  & 77.97 & 0      & 16.14 & 12409  & 74.52 & 0     & 32.89 & 48618 & 63.92 & 0       \\ 
FR-CNN$^*$~\cite{ren2017faster}& 9.01  & 4599  & 77.66 & 0      & 16.00 & 9477   & 74.17 & 0     & 33.11 & 37012 & 63.80 & 0       \\ 
PROSER~\cite{zhou2021learning} & 9.32  & 5105  & 77.35 & 7.48   & 16.65 & 10601  & 73.55 & 8.88  & 34.60 & 41569 & 63.09 & 11.15    \\
ORE~\cite{joseph2021towards}   & 8.39  & 4945  & 77.84 & 1.75   & 15.36 & 10568  & 74.34 & 1.81  & 32.40 & 40865 & 64.59 & 2.14    \\ 
DS~\cite{miller2018dropout}    & 8.30  & 4862  & 77.78 & 2.89   & 15.43 & 10136  & 73.67 & 4.11  & 31.79 & 39388 & 63.12 & 5.64    \\ 
\midrule
OpenDet       & \bf{6.44} & \bf{3944} & \bf{78.61} & \bf{9.05} & \bf{11.70} & \bf{8282}  & \bf{75.56} & \bf{12.30} & \bf{26.69} & \bf{32419} & \bf{65.55} & \bf{16.76} \\ 
\midrule
\end{tabular}%
}
    \vspace{-2mm}
    \caption{{\bf Comparisons with other methods on VOC-COCO-T$_2$.} Note that we put VOC-COCO-2$n$ in the appendix due to limited space.}
    \label{tab:voc_coco_b}
\end{table*}

\subsection{Unknown Probability Learner}
\label{subsec:upl}

As introduced in Sec.~\ref{subsec:cfl}, CFL expands low-density latent regions by narrowing the cluster of known classes (\emph{i.e.}, high-density regions).
However, we still lack explicit boundaries to separate high/low-density regions.
Traditional threshold-based methods with a small score threshold (\emph{e.g.}, 0.05) only maintain limited low-density regions, which cannot cover all unknown objects.
Here we present Unknown Probability Learner (UPL) to divide more low-density latent regions around the cluster of known classes.

To this aim, we first augment the K-way classifier with the K+1-way classifier, where K+1 denotes the unknown class.
Then the problem becomes: \emph{how to optimize the unknown class with only close-set training data?}
Let us consider a simple known \emph{vs.} unknown classifier with available open-set data, we can directly train a good classifier by maximizing margins between classes.
Now, we only have close-set data; To train such a classifier, we relax the maximum margin principle and only ensure all known objects are correctly classified, \emph{i.e.}, maintaining the close-set accuracy.
With this premise, we will introduce how to learn the unknown probability in the following section.

\noindent {\bf Review Cross-Entropy (CE) Loss.} We first review softmax CE Loss, the default classification loss of Faster R-CNN.
Let $\mathbf{s}$ denote the classification logits of a proposal, the softmax probability $\mathbf{p}$ of class $c$ is defined as:
\vspace{-1mm}
\begin{equation}
    p_c = {\rm softmax}(s_c) = \frac{ \exp (s_c)}{\sum_{j\in C} \exp (s_j)},
    \vspace{-1mm}
\end{equation}
where $C=C_{\mathcal{K} \cup \mathcal{U} \cup bg}$ denotes all known classes $C_\mathcal{K}$, unknown class $C_\mathcal{U}$ and background $C_{bg}$.
Then, we formulate softmax CE Loss $\mathcal{L}_{CE}$ as:
\vspace{-1mm}
\begin{equation} \label{eq:ce_loss}
    \mathcal{L}_{CE} = - \sum_{c \in C} y_c \log (p_c), 
    \quad
    y_c = \begin{cases}
        1, c=c^* \\
        0, c \neq c^*
        \end{cases},
    \vspace{-1mm}
\end{equation}
where $c^*$ means the ground truth class, and $y$ is the one-hot class label.
For simplicity, we re-write $\mathcal{L}_{CE}$ as:
\vspace{-1mm}
\begin{equation} \label{eq:ce_loss_short}
    \mathcal{L}_{CE} = - \log (p_{c^*}). 
    \vspace{-1mm}
\end{equation}

\noindent {\bf Learning Unknown Probability.} 
Since there is no supervision for the unknown probability $p_u$, we consider a conditional probability $p_u^{'}$ under the ground truth probability $p_{c^*}$.
Formally, we define $p_u^{'}$ as a softmax probability without the logit of ground truth class $c^*$:
\vspace{-1mm}
\begin{equation}
    p_u^{'} = \frac{ \exp (s_u)}{\sum_{j\in C, j \neq c^*} \exp (s_j)},
    \vspace{-1mm}
\end{equation}
where $u$ is short for the unknown class $C_\mathcal{U}$.
Then, similar to CE Loss, we formulate an Unknown Probability (UP) Loss $\mathcal{L}_{UP}$ to optimize $p_u^{'}$, which is defined as:
\begin{equation} \label{eq:up_loss}
    \mathcal{L}_{UP} = - \log (p_u^{'}).
\end{equation}
After that, we jointly optimize the CE Loss $\mathcal{L}_{CE}$ and UP Loss $\mathcal{L}_{UP}$ (illustrated in Fig.~\ref{fig:main_framework} (b)), where $\mathcal{L}_{CE}$ aims to maintain the close-set accuracy, and $\mathcal{L}_{UP}$ learns the unknown probability.
Take Fig.~\ref{fig:main_framework} (bottom-right) for an illustration, optimizing $\mathcal{L}_{UP}$ is equivalent to dividing more low-density latent regions (in gray color) from known classes.
Once we finished the training, the learned unknown probability serves as an indicator to identify unknown objects in these low-density regions.

\noindent {\bf Uncertainty-weighted Optimization.} 
Although we optimize the conditional probability $p_u^{'}$ instead of $p_u$, $\mathcal{L}_{UP}$ will still penalize the convergence of $\mathcal{L}_{CE}$, leading to the accuracy drop of known classes.
Inspired by uncertainty estimation~\cite{devries2018learning, sensoy2018evidential}, we add a weighting factor $w(\cdot)$ to $\mathcal{L}_{UP}$, which is defined as a function of $p_{c^*}$:
\begin{equation}\label{eq:upl_wx}
    w(p_{c^*}) = (1-p_{c^*})^\alpha p_{c^*},
\end{equation}
where $\alpha$ is a hyper-parameter ($\alpha$=1 by default).
Despite many design choices of $w(\cdot)$ (shown in Tab.~\ref{tab:upl_loss_wx}), we choose a simple yet effective one in Eq.~\ref{eq:upl_wx}.
We are inspired by the popular uncertainty signal: entropy $w(\mathbf{p})=-\mathbf{p}\log(\mathbf{p})$.
Since Eq.~\ref{eq:upl_wx} has a similar curve shape to entropy (see our appendix), it can also reflect uncertainty.
But our empirical findings suggest that Eq.~\ref{eq:upl_wx} is easier to optimize than entropy.
Finally, we formulate the uncertainty-weighted UP Loss as follow:
\begin{equation} \label{eq:wup_loss}
    \mathcal{L}_{UP} = - w(p_{c^*}) \log (p_u^{'}).
\end{equation}

\noindent {\bf Hard Example Mining.}
It is unreasonable to let all known objects learn the unknown probability as they do not belong to the unknown class.
Therefore, we present uncertainty-guided hard example mining to optimize $\mathcal{L}_{UP}$ with high-uncertainty proposals, which may overlap with real unknown objects in the latent space.
Here we consider two uncertainty-guided mining methods:
\begin{itemize}[leftmargin=*]
    \vspace{-2mm}
    \item \emph{Max entropy.} Entropy is a popular uncertainty measure~\cite{lakshminarayanan2016simple,malinin2018predictive} defined as: $H(\mathbf{p})=-\sum_{c \in C} p_c \log (p_c)$. For a mini-batch, We sort them in descending entropy order, and select \emph{top-k} examples.
    \vspace{-2mm}
    \item \emph{Min max-probability.} Max-probability, \emph{i.e.}, the maximum probability of all classes: $\max(\mathbf{p})$, is another uncertainty signal. We select \emph{top-k} examples with minimum max-probability.
    \vspace{-2mm}
\end{itemize}
Furthermore, since background proposals usually overwhelm the mini-batch, we sample the same number of foreground and background proposals, enabling our model to recall unknown objects from the background class.

\subsection{Overall Optimization}
\label{subsec:overall_optim}
Our method can be trained in an end-to-end manner with the following multi-task loss:
\vspace{-1mm}
\begin{equation}
    \mathcal{L} = \mathcal{L}_{rpn} + \mathcal{L}_{reg} + \mathcal{L}_{CE} + \beta \mathcal{L}_{UP} + \gamma_t \mathcal{L}_{IC},
    \vspace{-1mm}
\end{equation}
where $\mathcal{L}_{rpn}$ denotes the total loss of RPN, $\mathcal{L}_{reg}$ is smooth L1 loss for box regression, $\beta$ and $\gamma_t$ are weighting coefficients. Note $\gamma_t$ is proportional to the current iteration $t$ so that we can gradually decrease the weight of $\mathcal{L}_{IC}$ for better convergence of $\mathcal{L}_{CE}$ and $\mathcal{L}_{UP}$.

\section{Experiment}
\label{sec:exp}

\subsection{Experimental Setup}
\label{subsec:exp_setup}
\noindent {\bf Datasets.}
We construct an OSOD benchmark using popular PASCAL VOC~\cite{everingham2010pascal} and MS COCO~\cite{lin2014microsoft}.
We take the~\texttt{trainval} set of VOC for close-set training.
Meanwhile, we take 20 VOC classes and 60 non-VOC classes in COCO to evaluate our method under different open-set conditions.
Here we define two settings: {\bf VOC-COCO-\{T$_1$, T$_2$\}}. 
For setting {\bf T$_1$}, we gradually increase open-set classes to build three joint datasets with $n$=5000 VOC testing images and $\{n, 2n, 3n\}$ COCO images {\bf containing} \{20, 40, 60\} non-VOC classes, respectively.
For setting {\bf T$_2$}, we gradually increase the Wilderness Ratio (WR)~\footnote[2]{Wilderness Ratio is the ratio of \#images with unknown objects to \#images with known objects.}~\cite{dhamija2020overlooked} to construct four joint datasets with $n$ VOC testing images and $\{0.5n, n, 2n, 4n\}$ COCO images {\bf disjointing} with VOC classes.
See our appendix for more details.

\noindent {\bf Evaluation Metrics.} We use the {\bf Wilderness Impact (WI)}~\cite{dhamija2020overlooked} to measure the degree of unknown objects misclassified to known classes:
$WI = (\frac{P_\mathcal{K}}{P_{\mathcal{K} \cup \mathcal{U}}} - 1) \times 100$,
where $P_\mathcal{K}$ and $P_{\mathcal{K} \cup \mathcal{U}}$ denote the precision of close-set and open-set classes, respectively. Note that we scale the original WI by 100 for convenience. Following~\cite{joseph2021towards}, we report WI under a recall level of 0.8.
Besides, we also use {\bf Absolute Open-Set Error (AOSE)}~\cite{miller2018dropout} to count the number of misclassified unknown objects.
Furthermore, we report the {\bf mean Average Precision (mAP)} of known classes (\mAPK).
Lastly, we measure the novelty discovery ability by~\APU~(AP of the unknown class).
Note WI, AOSE, and~\APU~are open-set metrics, and \mAPK~is a close-set metric.

\noindent {\bf Comparison Methods.} We compare OpenDet with the following methods:
Faster R-CNN ({\bf FR-CNN})~\cite{ren2017faster}, Dropout Sampling ({\bf DS})~\cite{miller2018dropout}, {\bf ORE}~\cite{joseph2021towards} and {\bf PROSER}~\cite{zhou2021learning}.
FR-CNN is the base detector of other methods. We also report FR-CNN$^*$, which adopts a higher score threshold for testing.
We use the official code of ORE and reimplement DS and PROSER based on the FR-CNN framework.

\noindent {\bf Implementation Details.} We use ResNet-50~\cite{he2016resnet} with Feature Pyramid Network~\cite{lin2017feature} as the backbone of all methods. 
We adopt the same learning rate schedules with Detectron2~\cite{wu2019detectron2}. 
SGD optimizer is adopted with an initial learning rate of 0.02, momentum of 0.9, and weight decay of 0.0001.
All models are trained on 8 GPUs with a batch size of 16. 
For {\bf CFL}, we set memory size $Q$=256 and sampling size $q$=16. We sample proposals with an IoU threshold $T_m$=0.7 for the memory bank, and $T_b$=0.5 for the mini-batch. 
For {\bf UPL}, we sample $k$=3 examples for foreground and background proposals respectively. Besides, we set hyper-parameters $\alpha$=1.0 and $\beta$=0.5. We set the initial value of $\gamma_t$=0.1 and linearly decrease it to zero.

\begin{table}[t]
    \centering
\small
\setlength{\tabcolsep}{2.8mm}{
\begin{tabular}{cc|cccc}
\hline
CFL & UPL & WI$_\downarrow$   & AOSE$_\downarrow$  & mAP$_\mathcal{K \uparrow}$   & AP$_\mathcal{U \uparrow}$ \\
\hline
\multicolumn{2}{c|}{\gray{baseline}}& \gray{19.26} & \gray{16433} & \gray{58.33} & \gray{0}   \\
$\checkmark$ &              & 17.92 & 15162 & 58.54 & 0     \\
             & $\checkmark$ & 16.47 & 12018 & 57.91 & 14.27 \\
$\checkmark$ & $\checkmark$ & {\bf 14.95} & {\bf 11286} & {\bf 58.75} & {\bf 14.93} \\
\hline
\end{tabular}%
}
    \caption{{\bf Effect of different components} on VOC-COCO-20.}
    \vspace{-3mm}
    \label{tab:overall_improve}
    \end{table}

\subsection{Main Results}
We compare OpenDet with other methods on VOC-COCO-\{T$_1$, T$_2$\}. 
Tab.~\ref{tab:voc_coco_a} shows results on VOC-COCO-T$_1$ by gradually increasing unknown classes.
Compared with FR-CNN, FR-CNN$^*$ with a higher score threshold (0.05$\rightarrow$0.1) does not reduce WI, but results in a decrease in~\mAPK, where known objects with low confidence are filtered out.
PROSER improves AOSE and \APU~to some extent, but the WI and~\mAPK~are even worse. Although ORE and DS achieve comparable \mAPK, the improvement on open-set metrics is limited. 
The proposed OpenDet outperforms other methods by a large margin. Taking VOC-COCO-20 for an example, OpenDet gains about 20\%, 25\%, 14.93 on WI, AOSE and~\APU~respectively without compromising the~\mAPK~(58.75 \emph{vs.} 58.45).
Besides, we also report~\mAPK~on VOC, which indicates OpenDet is competitive in the traditional close-set setting (80.02 \emph{vs.} 80.10).

We also compare OpenDet with other methods by increasing the WR, where the results in Tab.~\ref{tab:voc_coco_b} draws similar conclusions with Tab.~\ref{tab:voc_coco_a}.
Our method performs better as the WR increases. For example, the~\mAPK~gains on VOC-COCO-\{0.5$n$, $n$, 4$n$\} are \{0.64, 1.04, 1.63\}, indicating that our method actually separates known and unknown classes.

\subsection{Ablation Studies}
\label{subsec:ablation}
In this section, we conduct ablation experiments on VOC-COCO-20 to analyze the effect of our main components and core design choices.

\noindent {\bf Overall Analysis.} We first analyze the contribution of different components.
As shown in Tab.~\ref{tab:overall_improve}, our two modules, CFL and UPL, show substantial improvement compared with the baseline.
The combination of CFL and UPL further boosts the performance. 
We also visualize the latent features in Fig.~\ref{fig:tsne_vis}, where our method learns clear separation between known and unknown classes.

\begin{table}[t]
    \centering
\small
\begin{tabular}{l|c|cc}
\hline
Memory                              & size   & WI$_\downarrow$  & mAP$_\mathcal{K \uparrow}$ \\
\hline
single-GPU mini-batch               & \~{}50          & 16.19 & 58.29   \\
cross-GPU  mini-batch               & \~{}50$\times$8 & 15.88 & 58.07   \\
class-agnostic memory bank          & 5120            & 15.99 & 57.47   \\
class-agnostic memory bank$^*$      & 65536           & 15.49 & {\bf 58.90}  \\
class-balanced memory bank          & 256$\times$20   & {\bf 14.95} & 58.75   \\
\hline
\end{tabular}%


    \caption{{\bf Class-balanced memory bank.} We compare our class-balanced memory bank with other variants.
    We keep class-agnostic memory bank the same size with ours (256$\times$20=5120). 8 and 20 are the number of GPU and VOC classes, respectively. $^*$ means a larger memory size. }
    \vspace{-2mm}
    \label{tab:cfl_memory_bank}
    \end{table}

\begin{table}[t]
    \centering
\small
\subfloat[IoU threshold]{\label{tab:cfl_memory_iou}
    \resizebox{\linewidth}{!}{%
    \begin{tabular}{c|cccccc}
    \hline
                                & {\bf (a)} & {\bf (b)} & {\bf (c)} & {\bf (d)} & {\bf (e)} & {\bf (f)}  \\
    T$_b$                       & 0.5 & 0.7 & 0.9 & 0.5 & 0.5 & 0.7  \\
    T$_m$                       & 0.5 & 0.7 & 0.9 & 0.7 & 0.9 & 0.9  \\
    \hline            
    WI$_\downarrow$             &15.33&15.16&15.27&14.95&{\bf 14.62}&15.27 \\
    mAP$_\mathcal{K \uparrow}$  &58.29&58.55&58.32&{\bf 58.75}&58.66&58.33 \\
    \hline
    \end{tabular}%
    }
}
\\
\vspace{-3mm}
\subfloat[Memory size and mini-batch sampling size]{\label{tab:cfl_memory_size}
    \resizebox{\linewidth}{!}{%
    \begin{tabular}{c|cccccc}
    \hline
                                & {\bf (a)} & {\bf (b)} & {\bf (c)} & {\bf (d)} & {\bf (e)} & {\bf (f)}  \\
    q                          & 16    &  16   &  16   &  32    &  64   &  128 \\
    Q                          & 128   &  256  & 512   & 256    &  256  &  256 \\
    \hline
    WI$_\downarrow$            & 15.36 & 14.95 & {\bf 14.47} & 15.24  & 15.43 & 14.77 \\
    mAP$_\mathcal{K \uparrow}$ & 58.51 & {\bf 58.75} & 58.31 & 58.32  & 57.77 & 58.18 \\
    \hline
    \end{tabular}%
    }
}

    \caption{{\bf Sampling strategy in CFL}. We list different choices of {\bf(a)} memory sampling threshold T$_m$ and mini-batch sampling threshold T$_b$, {\bf (b)} memory size $Q$ and sampling size $q$.}
    \vspace{-2mm}
    \label{tab:cfl_memory_sampling}
    \end{table}

\noindent {\bf Contrastive Feature Learner.} We carefully study the design choices of the memory bank and example sampling strategy in CFL.
As $\mathcal{L}_{IC}$ is optimized between the current mini-batch and the memory bank, we investigate different {\bf designs of memory} in Tab.~\ref{tab:cfl_memory_bank}.
Compared with the mini-batch (\emph{i.e.}, short-term memory), the settings with a memory bank perform better on WI.
However, imbalanced training data makes the class-agnostic memory bank filled with high-frequency classes, leading to a drop in \mAPK~(58.76$\rightarrow$57.47).
Enlarging the memory bank size (5120$\rightarrow$65536) can alleviate this issue, but it requires more computation.
The proposed class-balanced memory bank can store more diverse examples with a small memory size, outperforming other variants.

We further study the design choices of {\bf example sampling strategies}. For a mini-batch, we consider the IoU threshold T$_b$; for the memory bank, we consider the IoU threshold T$_m$, memory size $Q$ and mini-batch sampling size $q$.
As shown in Tab.~\ref{tab:cfl_memory_iou}, the settings (d) and (e) achieve the best result in \mAPK~and WI, respectively, while (a)-(c) are worse than (d)-(e) in WI.
This indicates that the mini-batch requires a loose constraint to gather more diverse examples, while the memory bank needs high-quality examples to represent the class centers.
In Tab.~\ref{tab:cfl_memory_size}, (b) and (c) perform better than other settings, which demonstrates that long-term memory (\emph{i.e.}, larger $Q/q$) is a good choice for CFL.

\begin{table}[t]
    \centering
\small
\begin{tabular}{c|c|cccc}
\hline
& $w(\cdot)$                        & WI$_\downarrow$  & AOSE$_\downarrow$  & mAP$_\mathcal{K \uparrow}$ & AP$_\mathcal{U \uparrow}$   \\
\hline
              & \gray{baseline}                      & \gray{19.26} & \gray{16433} & \gray{58.33} & \gray{0}   \\
\textbf{(a)}  & identity                             & {\bf 10.50} & 12185       & 56.42       & 11.33           \\
\textbf{(b)}  & $-p_{c^*}\log(p_{c^*})$              & 14.70       & 11384       & 58.13       & 13.71           \\
\textbf{(c)}  & $(1-p_{c^*})^\alpha p_{c^*}$         & 14.95       & {\bf 11286} & {\bf 58.75} & 14.93           \\
\hline
\textbf{(d)}  & $(1-p_m)^\alpha p_m$                 & 14.86       & 11296       & 58.03       & 14.15           \\
\textbf{(e)}  & $ H(\mathbf{p})/ \log(C)$            & 14.29   & 11690       & 57.75       & 14.65           \\
\hline
\end{tabular}%
    \caption{{\bf Different designs of $w(\cdot)$ in $\mathcal{L}_{UP}$.}
    $p_m$ is the maximum probability of all classes: $p_m={\rm max}(\mathbf{p})$. {\bf (e)} denotes normalized entropy where $H(\mathbf{p})=-\sum_c p_c \log(p_c)$ and $C$ is the number of known classes.
    }
    \vspace{-2mm}
    \label{tab:upl_loss_wx}
    \end{table}

\begin{table}[t]
    \centering
\small
\begin{tabular}{l|cccc}
\hline
Setting       & WI$_\downarrow$   & AOSE$_\downarrow$  & mAP$_\mathcal{K \uparrow}$   & AP$_\mathcal{U \uparrow}$ \\
\hline
\textbf{OpenDet} (\emph{w/} HEM)         & 14.95 & {\bf 11286} & {\bf 58.75} & {\bf 14.93} \\
\textbf{(a)} \emph{w/o} HEM       & 18.33 & 13733 & 57.41 & 13.91 \\
\textbf{(b)} \emph{w/o} bg.       & {\bf 13.02} & 12230 & 56.53 & 13.49 \\
\hline
\textbf{(c)} top-k: \\
1     & {\bf 14.46} & 12826       & 58.42       & 14.54  \\
3     & 14.95       & 11286       & {\bf 58.75} & {\bf 14.93}  \\
5     & 14.66       & 10412       & 58.50       & 14.55  \\
10    & 15.15       & {\bf 10358} & 58.25       & 14.86  \\
all   & 18.40       & 11779       & 56.55       & 13.89  \\
\hline
\textbf{(d)} metric: \\
random        & 17.01 & 13065 & 56.99 & {\bf 15.58} \\
max entropy   & {\bf 14.29} & 11514 & 58.27 & 15.46 \\
min max-probability & 14.95 & {\bf 11286} & {\bf 58.75} & 14.93 \\
\hline
\end{tabular}%
    \caption{{\bf Hard example mining (HEM) in UPL.}
    {\bf (a)} without HEM. {\bf (b)} without background: we only sample foreground proposals. {\bf (c)} varying \emph{top-k}. the setting \emph{all} means all foreground and equal number of background proposals. {\bf (d)} mining methods.
    }
    \vspace{-2mm}
    \label{tab:upl_hard_example_mining}
    \end{table}

\begin{figure*}[!t]
        \centering
        \includegraphics[width=0.98\linewidth]{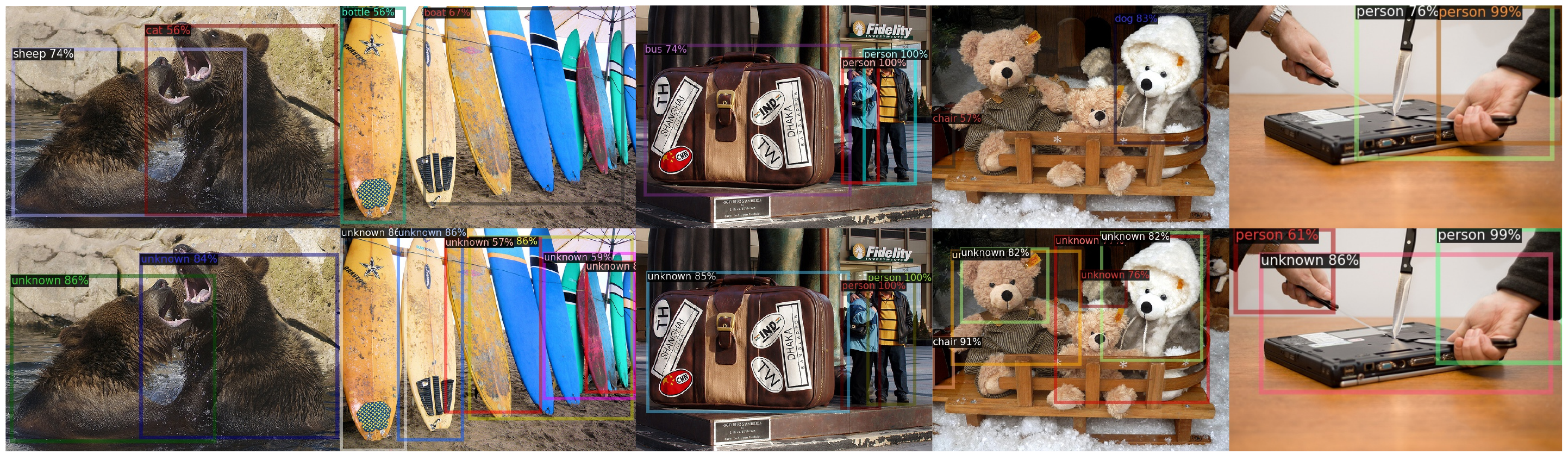}
        \caption{{\bf Qualitative comparisons} between the baseline (top) and OpenDet (bottom).
        We train both models on VOC and visualize the detection results on COCO. Note that we apply NMS between known classes and the unknown class for better visualization.}
        \label{fig:qualitative_results}
    \end{figure*}

\noindent {\bf Unknown Probability Learner.}
We first explore {\bf different variants of $w(\cdot)$}.
Compared with the baseline, Tab.~\ref{tab:upl_loss_wx} (a) significantly reduces WI and AOSE, but leads to \mAPK~drop, which indicates the learned unknown probability is overestimated.
The formula of (b) is similar to entropy, and (c) is our default setting.
As discussed in Sec.~\ref{subsec:upl}, both (b) and (c) achieve satisfactory results in WI and AOSE, but (c) is outperforms (b) in \mAPK~and \APU.
(d) and (e) are two variants of $w(\cdot)$ based on max-probability and entropy, respectively. They obtain comparable performance on open-set metrics, but the \mAPK~is lower than (c).

We also analyze the effect of {\bf Hard example mining (HEM)} in Tab.~\ref{tab:upl_hard_example_mining}.
Comparing Tab.~\ref{tab:upl_hard_example_mining} (a) (without HEM) with our default setting (with HEM), we show HEM is crucial for UPL.
Tab.~\ref{tab:upl_hard_example_mining} (b) indicates background proposals are also necessary for unknown probability learning, \emph{e.g.}, OpenDet without background leads to 2.22 and 1.44 drop in~\mAPK~and~\APU.
Besides, we varying the hyper-parameter \emph{top-k} in Tab.~\ref{tab:upl_hard_example_mining} (c) where HEM works in a wide range of $k$ ($\leq$10), while optimizing all examples is not applicable.
Tab.~\ref{tab:upl_hard_example_mining} (d) demonstrates the effectiveness of two mining methods, \emph{i.e.}, max entropy and min max-probability. 

\noindent {\bf Qualitative Comparisons.} Fig.~\ref{fig:qualitative_results} compares the qualitative results of baseline and OpenDet. OpenDet gives an \texttt{unknown} label to unknown objects (bottom row), while the baseline method classifies them to known classes or the background (top row). See our appendix for more qualitative results.
\vspace{-0.5mm}

\subsection{Extend to One-Stage Detector}
Although OpenDet is based on a two-stage detector, it can be easily extended to other architectures, \emph{e.g.}, a representative one-stage detector RetinaNet~\cite{lin2017focal}.
RetinaNet has a backbone and two parallel sub-networks for classification and regression, respectively.
Different from FR-CNN, RetinaNet adopts Focal Loss~\cite{lin2017focal} for dense classification.
Here we show how to extend OpenDet to RetinaNet (denote with Open-RetinaNet).
For CFL, we append the contrastive head to the second-last layer of the classification sub-network. We adopt the same sampling strategies in CFL and optimize $\mathcal{L}_{IC}$ with pixel-wise features.
For UPL, we only sample hard foreground examples as RetinaNet does not preserve a background class. Then $\mathcal{L}_{UP}$ is jointly optimized with Focal Loss.
Tab.~\ref{tab:open_retinanet} reports the results on VOC-COCO-20, where Open-RetinaNet shows significant improvements on all open-set metrics and achieves comparable close-set \mAPK. For example, Open-RetinaNet gains 23.7\%, 55.8\%, and 11.02 in WI, AOSE, and \APU, respectively.
\begin{table}
    \addtolength{\tabcolsep}{0.5mm}
\begin{center}
\small
\begin{tabular}{l|cccc}
\hline
Method            & WI$_\downarrow$   & AOSE$_\downarrow$  & mAP$_\mathcal{K \uparrow}$   & AP$_\mathcal{U \uparrow}$ \\
\hline
RetinaNet         & 14.58 & 38071 & 57.44 & 0    \\
Open-RetinaNet & {\bf 10.84} & {\bf 16815} & 57.25 & {\bf 11.02} \\
\hline
\end{tabular}%
\end{center}

    \vspace{-3mm}
    \caption{{\bf Performance of Open-RetinaNet} on VOC-COCO-20.}
    \label{tab:open_retinanet}
\end{table}

\section{Conclusions}
This paper proposes a novel Open-set Detector (OpenDet) to solve the challenging OSOD task by expanding low-density latent regions.
OpenDet consists of two well-designed learners, CFL and UPL, where CFL performs \emph{instance-level} contrastive learning to learn more compact features and UPL learns the unknown probability that serves as a threshold to further separate known and unknown classes.
We also build an OSOD benchmark and conduct extensive experiments to demonstrate the effectiveness of our method.
Compared with other methods, OpenDet shows significant improvements on all metrics.

\noindent {\bf Limitations.}
We notice that some low-quality proposals belonging to known classes are given the \texttt{unknown} label during inference, and cannot be filtered out by per-class non-maximum suppression.
Although these proposals do not hurt the close-set~\mAPK, it raises a new question about reducing false unknown predictions, which is also a direction for our future work.

\section*{Acknowledgement}
This work was supported by National Nature Science Foundation of China under grant 61922065, 41820104006 and 61871299. The
numerical calculations in this paper have been done on the supercomputing system in the Supercomputing Center of Wuhan University. Jian Ding is also supported by China Scholarship Council.
\appendix
\setcounter{table}{0}
\renewcommand{\thetable}{A\arabic{table}}
\setcounter{figure}{0}
\renewcommand\thefigure{A\arabic{figure}}

\section{More Experimental Details}

\subsection{Datasets}
In this section, we introduce more details about the dataset construction.

\noindent {\bf PASCAL VOC~\cite{everingham2010pascal}.} We use VOC07 \texttt{train} and VOC12 \texttt{trainval} splits for the training, and VOC07 \texttt{test} split to evaluate the close-set performance.
We take VOC07 \texttt{val} as the validation set.

\noindent {\bf VOC-COCO-T$_1$.} We divide 80 COCO classes into four groups (20 classes per group) by their semantics~\cite{joseph2021towards}: 
{\bf (1)} VOC classes. {\bf (2)} Outdoor, Accessories, Appliance, Truck. {\bf (3)} Sports, Food. {\bf (4)} Electronic, Indoor, Kitchen, Furniture.
We construct VOC-COCO-\{20, 40, 60\} with $n$=5000 VOC testing images and \{$n$, 2$n$, 3$n$\} COCO images {\bf containing} \{20, 40, 60\} non-VOC classes with semantic shifts, respectively.
Note that we only ensure each COCO image contains objects of corresponding open-set classes, which means objects of VOC classes will also appear in these images.
This setting is more similar to real-world scenarios where detectors need to carefully identify unknown objects and do not classify known objects into the unknown class.

\noindent {\bf VOC-COCO-T$_2$.} We gradually increase the Wilderness Ratio to build four dataset with $n$=5000 VOC testing images and \{0.5$n$, $n$, 2$n$, 4$n$\} COCO images {\bf disjointing} with VOC classes.
Compared with the setting {\bf T$_1$}, {\bf T$_2$} aims to evaluate the model under a higher wilderness, where large amounts of testing instances are not seen in the training.

\noindent {\bf Comparisons with existing benchmarks.}
\cite{dhamija2020overlooked} proposed the first OSOD benchmark. They also use the data in VOC for close-set training, and both VOC and COCO for open-set testing. In the testing phase, they just vary the number of open-set images sampled from COCO, while ignoring the number of open-set categories.
~\cite{joseph2021towards} proposed an open world object detection benchmark. They divide the open-set testing set into several groups by category. However, the wilderness ratio of each group is limited, and such data partitioning cannot reflect the real performance of detectors under extreme open-set conditions.
In contrast, our proposed benchmark considers both the number of open-set classes (VOC-COCO-T$_1$) and images (VOC-COCO-T$_2$).

On the other hand, some works on open-set panoptic segmentation~\cite{hwang2021exemplar} divide a single dataset into close-set and open-set. If a image contains both close-set and open-set instances, they just remove the annotations of open-set instances.
Differently, we strictly follows the definition in OSR~\cite{scheirer2012toward} that unknown instances should not appear in training.
To acquire enough open-set examples, we take both VOC and COCO from cross-dataset evaluation, which is a common practice in OSR~\cite{kong2021opengan,sun2020conditional,zhou2021learning}.

\subsection{Implementation Details}
\noindent {\bf Training schedule.} Inspired by~\cite{vaze2021open} that a good close-set classifier benefits OSR, we train all models with the 3$\times$ schedule (\emph{i.e.}, 36 epochs). 
Besides, we enable UPL after several warmup iterations (\emph{e.g.}, 100 iterations) to make sure the model produce valid probabilities.

\noindent {\bf Open-RetinaNet.} We change some hyper-parameters for Open-RetinaNet.
In OpenDet, we take object proposals as examples and apply CFL to proposal-wise embeddings, which are equivalent to the anchor boxes in RetinaNet.
Therefore, we optimize Instance Contrastive Loss $\mathcal{L}_{IC}$ with pixel-wise features of each anchor box.
Since the number of anchor box is much larger than the proposals in OpenDet, we enlarge the memory size $Q$=1024, sampling size $q$=64, and loss weight to 0.2 in CFL.
Similar, we sample 10 hard examples rather than 3 in UPL.

\subsection{Evaluation Metrics}
Firstly, we give a detailed formulation of the Wilderness Impact~\cite{dhamija2020overlooked}, which is defined as:
\begin{equation}
\begin{aligned}
    WI &= \frac{P_\mathcal{K}}{P_\mathcal{K \cup U}} - 1  \\
       &=  \frac{TP_\mathcal{K}}{TP_\mathcal{K}+FP_\mathcal{K}} / \frac{TP_\mathcal{K}}{TP_\mathcal{K}+FP_\mathcal{K}+FP_\mathcal{U}}  - 1 \\
       &= \frac{FP_\mathcal{U}}{TP_\mathcal{K}+FP_\mathcal{K}},
\end{aligned}
\end{equation}
where $FP_\mathcal{U}$ means that any detections belonging to the unknown classes $C_\mathcal{U}$ are classified to one of known classes $C_\mathcal{K}$.
For \APU~(AP of unknown classes), we merge the annotations of all unknown classes into one class, and calculate the \emph{class-agnostic} AP between unknown's predictions and the ground truth.

\section{Additional Main Results}
Due to limited space in our main paper, we report the results on VOC-COCO-2$n$ in Tab.~\ref{tab:voc_coco_2n}, where OpenDet shows significant improves than other methods.

\begin{table}[h]
    \centering
\small
\setlength{\tabcolsep}{2.8mm}{
\begin{tabular}{l|cccc}
\midrule
Method                         & WI$_\downarrow$   & AOSE$_\downarrow$  & mAP$_\mathcal{K \uparrow}$   & AP$_\mathcal{U \uparrow}$  \\ \midrule
FR-CNN~\cite{ren2017faster}    & 24.18  & 24636  & 70.07 & 0    \\    
FR-CNN$^*$~\cite{ren2017faster}& 24.05  & 18740  & 69.81 & 0    \\    
PROSER~\cite{zhou2021learning} & 25.74  & 21107  & 69.32 & 10.31 \\   
ORE~\cite{joseph2021towards}   & 23.67  & 20839  & 70.01 & 2.13 \\  
DS~\cite{miller2018dropout}    & 23.21  & 20018  & 69.33 & 4.84 \\  
\midrule
OpenDet       & \bf{18.69} & \bf{16329} & \bf{71.44} & \bf{14.96} \\
\midrule
\end{tabular}%
}
    \caption{{\bf Comparisons with other methods on VOC-COCO-2$n$.}
    This table is an extension of Tab.2 in our main paper.
    }
    \label{tab:voc_coco_2n}
    \end{table}

\section{Additional Ablation Studies}

\begin{figure}[t]
    \includegraphics[width=\linewidth]{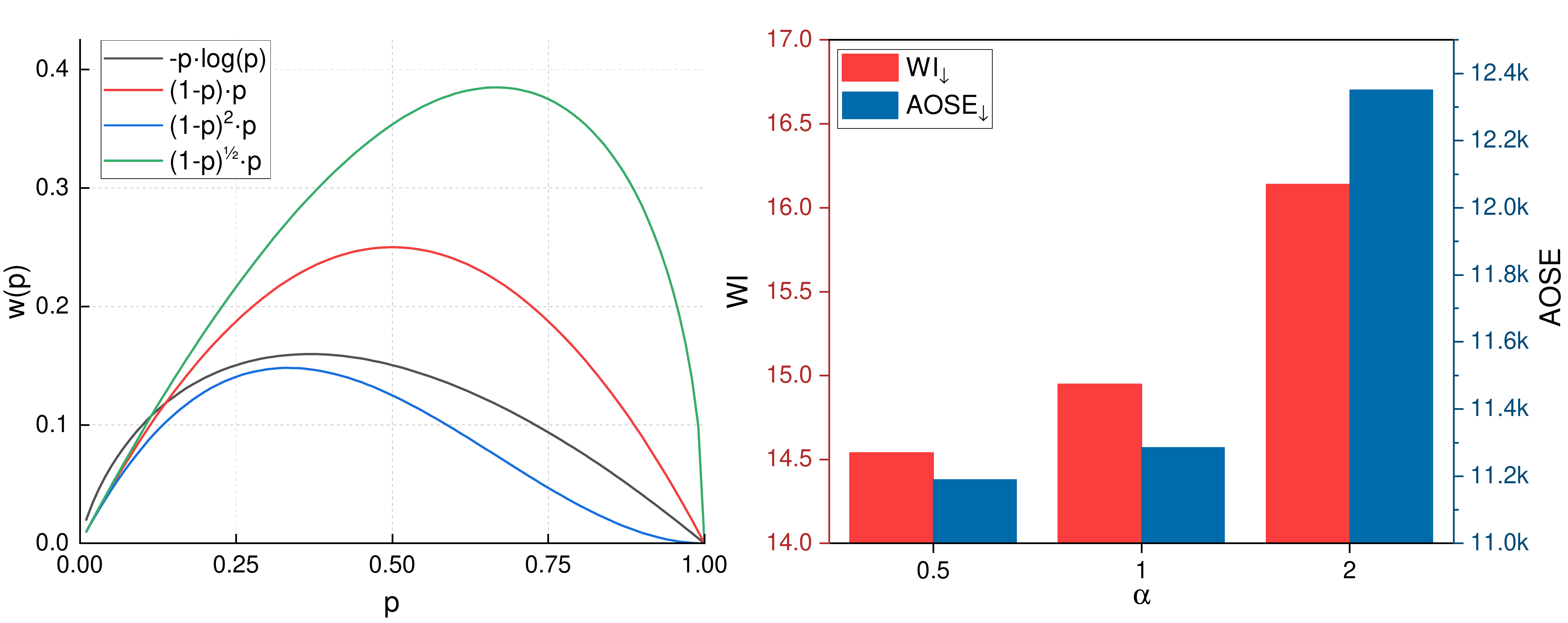}
    \vspace{-3mm}
    \caption{{\bf Visualization of different $w(\cdot)$.}}
    \label{fig:plot_wup}
\end{figure}
    
\begin{table}[t]
    \small
\centering
\addtolength{\tabcolsep}{1mm}
\begin{tabular}{c|cccc}
\hline
metric         & baseline   & +CFL       & +UPL       & Ours       \\
\hline
intra-variance & 3.79  & 2.83  & 3.05  & {\bf 2.47}  \\
inter-distance & 62.74 & 65.17 & 64.69 & {\bf 66.31} \\
\hline
\end{tabular}%

    \caption{{\bf Quantitative analyses of the latent space.} We calculate the intra-class variance and inter-class distance of latent features.}
    \label{tab:region_expanding}
\end{table}

\begin{table}[t]
    \begin{center}
\small
\setlength{\tabcolsep}{2.6mm}{
\begin{tabular}{c|ccccc}
\hline
$\gamma_t$                 & 0.01  &  {\bf 0.1}  & 0.5 & 1.0 & w/o decay   \\
\hline
WI$_\downarrow$            & 16.13 & 14.95 & 12.26 & 9.71  & 15.65 \\
mAP$_\mathcal{K \uparrow}$ & 58.90 & 58.75 & 57.47 & 53.36 & 58.43 \\
\hline
\end{tabular}%
}
\end{center}

    \vspace{-3mm}
    \caption{{\bf Loss weight of $\mathcal{L}_{IC}$.} \emph{w/o} decay: $\gamma_t$ is a constant (\emph{i.e.}, 0.1) instead of variable.}
    \label{tab:weight_cfl}
    \end{table}

\begin{table}[t]
    \addtolength{\tabcolsep}{3mm}
\centering
\small
\begin{tabular}{c|ccc}
\hline
$\tau$                     & 0.07~\cite{he2020momentum}   &  {\bf 0.1}~\cite{khosla2020supervised}  & 0.2    \\
\hline
WI$_\downarrow$            & 15.48  & 14.95  & 15.50  \\
mAP$_\mathcal{K \uparrow}$ & 57.80  & 58.75  & 58.87  \\
\hline
\end{tabular}%
    \caption{{\bf Temperature $\tau$ in $\mathcal{L}_{IC}$.}}
    \label{tab:tau_cfl}
    \end{table}

\noindent {\bf Visual analyses of $w(\cdot)$.} 
In Fig.~\ref{fig:plot_wup}, we plot the graph of different $w(\cdot)$.
Compared with entropy: $-p\log(p)$, the proposed function $(1-p)^\alpha \cdot p$ can adjsut the curve shape by changing $\alpha$.
In other words, the model adjusts the weights of examples as $\alpha$ changes.
The right of Fig.~\ref{fig:plot_wup} reports the model's open-set performance by varying $\alpha$, where smaller $\alpha$ reduces WI and AOSE.

\noindent {\bf Quantitative analyses of latent space.}
In Fig.~\ref{fig:tsne_vis} of the main paper, we give a visual analyses of latent space. Here we give a quantitative analyses of latent space in Tab.~\ref{tab:region_expanding}.
Specifically, we calculate the intra-class variance and inter-class distance of latent features.
Tab.~\ref{tab:region_expanding} shows that CFL and UPL, as well as their combination reduce intra-class variance and enlarge inter-class distance.
The results further confirm our conclusion in the main paper that our method can expand low-density latent regions.

\noindent {\bf More hyper-parameters in CFL.}
{\bf Loss weight:} Tab.~\ref{tab:weight_cfl} shows that loss weight is important for $\mathcal{L}_{IC}$, where a small weight (\emph{e.g.}, 0.01) cannot learn compact features and a large weight (\emph{e.g.}, 1.0) hinder the generalization ability.
Besides, Tab.~\ref{tab:weight_cfl} (last column) also demonstrates the effectiveness of loss decay.
{\bf Temperature:} We try different $\tau$ that used in pervious works~\cite{he2020momentum,khosla2020supervised}.
Tab.~\ref{tab:tau_cfl} indicates that $\tau$=0.1~\cite{khosla2020supervised} works better than other settings.

\noindent {\bf Training strategy.}
Some works in OSR~\cite{zhou2021learning} adopted a pretrain-then-finetune paradigm to train the unknown identifier.
We carefully design the UPL so that OpenDet can be trained in an end-to-end manner.
Tab.~\ref{tab:e2e_vs_finetune} shows that jointly optimizing UPL performs better than that of fine-tuning.

\noindent {\bf Open-RetinaNet.}
To further demonstrates the effectiveness of Open-RetinaNet,
we report more results in Tab.~\ref{tab:open_retinanet_more},
where Open-RetinaNet shows substantial improvements on WI, AOSE and~\APU, and achieves comparable performance on~\mAPK.

\noindent {\bf Vision transformer as backbone.}
We find the detector with vision transformer, \emph{e.g.}, Swin Transformer~\cite{liu2021swin} is a stronger baseline for OSOD.
As shown in Tab.~\ref{tab:swin_backbone}, models with a Swin-T backbone significantly suppress their ResNet counterparts.

\noindent {\bf Speed and computation.}
In the training stage, OpenDet only increases 14\% (1.4h \emph{vs.} 1.2h) training time and 1.2\% (2424Mb \emph{vs.} 2395Mb) memory usage.
In the testing phase, as we only add the unknown class to the classifier, OpenDet keeps similar running speed and computation with FR-CNN.

\begin{table}[t]
    \centering
\small
\setlength{\tabcolsep}{3.2mm}{
\begin{tabular}{l|cccc}
\hline
setting        & backbone & epoch & WI$_\downarrow$  & mAP$_\mathcal{K \uparrow}$    \\
\hline
end-to-end                  & -          & -     & {\bf 14.95} & {\bf 58.75}           \\
\hline
\multirow{3}{*}{fine-tune}  & fixed      & 1     & 17.98 & 56.88   \\
                            & fixed      & 12    & 17.43 & 56.86   \\
                            & trainable  & 12    & 17.01 & 57.19   \\
\hline
\end{tabular}%
}
    \caption{{\bf End-to-end \emph{vs.} fine-tune in UPL.}
    {\bf End-to-end:} we jointly optimize UPL and other modules in OpenDet. 
    {\bf Fine-tune:} we pretrain a model without UPL, and optimize UPL in the fine-tuning stage.
    }
    \label{tab:e2e_vs_finetune}
    \end{table}

\begin{table}[t]
    \centering
\small
\setlength{\tabcolsep}{2.8mm}{
\begin{tabular}{l|cccc}
\hline
 Method  & WI$_\downarrow$  & AOSE$_\downarrow$  & mAP$_\mathcal{K \uparrow}$ & AP$_\mathcal{U \uparrow}$   \\
 \hline
 \emph{VOC:} \\
 RetinaNet         &  -  &  -    &  {\bf 79.84}   & -   \\
 Open-RetinaNet    &  -  &  -    &  79.72   & -   \\
 \hline
\emph{VOC-COCO-40:} \\
RetinaNet         & 17.60   &  58383  &  {\bf 53.81}   & 0    \\
Open-RetinaNet    & {\bf 13.65}   &  {\bf 25964}  &  53.22   & {\bf 8.23} \\
\hline
\emph{VOC-COCO-60:} \\
RetinaNet         & 14.20   & 64327   & {\bf 54.68}    & 0    \\
Open-RetinaNet    & {\bf 11.28}   & {\bf 30631}   & 54.25    & {\bf 3.20} \\
\hline
\end{tabular}%
}
    \vspace{-1mm}
    \caption{{\bf Open-RetinaNet on more datasets.}}
    \label{tab:open_retinanet_more}
    \end{table}

\begin{table}[t]
    \centering
\small
\resizebox{\linewidth}{!}{%
\begin{tabular}{c|c|cccc}
\hline
Method                   & backbone  & WI$_\downarrow$  & AOSE$_\downarrow$  & mAP$_\mathcal{K \uparrow}$ & AP$_\mathcal{U \uparrow}$   \\
\hline
\multirow{2}{*}{FR-CNN}  & ResNet-50 & 18.39 & 15118 & 58.45 & 0 \\
                         & {\bf Swin-T}    & {\bf 15.99} & {\bf 13204} & {\bf 63.09} & 0   \\
\hline
\multirow{2}{*}{OpenDet} & ResNet-50 & 14.95 & 11286 & 58.75 & 14.93 \\
                         & {\bf Swin-T}    & {\bf 12.51} & {\bf 9875}  & {\bf 63.17} & {\bf 15.77} \\
\hline
\end{tabular}%
}
    \vspace{-1mm}
    \caption{{\bf Comparisons of different backbones}, \emph{i.e.}, ResNet-50~\cite{he2016resnet} and Swin-T~\cite{liu2021swin}.}
    \label{tab:swin_backbone}
    \end{table}

\section{Comparison with ORE~\cite{joseph2021towards}}
\noindent {\bf Implementation details.}
The original ORE adopted a R50-C4 FR-CNN framework, and train the model with 8 epochs.
For fair comparisons, we replace the R50-C4 architecture with R50-FPN, and train all models with 3$\times$ schedule.
Besides, as discussed in these issues\footnote[1]{\href{https://github.com/JosephKJ/OWOD/issues?q=is:issue+cannot+reproduce}{https://github.com/JosephKJ/OWOD/issues?q=cannot+reproduce}},
we report our re-implemented results when comparing with ORE in an open world object detection task (see Tab.~\ref{tab:owod_t1}).

\noindent {\bf Analysis of ORE.} To learn the energy-based unknown identifier (Sec 4.3 in~\cite{joseph2021towards}), ORE requires an additional validation set with the annotations of unknown classes.
We notice that ORE continues to train on the validation set, so that the model can leverage the information of unknown classes.
In Tab.~\ref{tab:ore}, we find ORE without training on valset (\emph{i.e.}, froze parameters) obtains a rather lower \mAPK~(53.96 \emph{vs.} 58.45), and large amounts of known examples are misclassified to unknown.
In contrast, OpenDet outperforms ORE without using the information of unknown classes.

\begin{table}[t]
    \centering
\small
\resizebox{\linewidth}{!}{%
\begin{tabular}{c|c|cccc}
\hline
Method & \makecell[c]{train model \\ on valset}  & WI$_\downarrow$   & AOSE$_\downarrow$  & mAP$_\mathcal{K \uparrow}$   & AP$_\mathcal{U \uparrow}$  \\
\hline
FR-CNN  & $\times$ & 18.39 & 15118 & 58.45 & 0     \\
\hline
\multirow{2}{*}{ORE} &
$\times$            & 8.46      & 2909     & \textcolor{red}{53.96}     & 9.64  \\
& \cg $\checkmark$  & \cg 16.98 & \cg 12868 & \cg 58.35 & \cg 5.13 \\
\hline
OpenDet & $\times$ & 14.95 & 11286 & 58.75 & 14.93 \\
\hline
\end{tabular}%
}
    \vspace{-1mm}
    \caption{{\bf Comparison with ORE~\cite{joseph2021towards}.}
    The row with \textcolor{gray}{gray} background is reported in our main paper.
    }
    \label{tab:ore}
    \end{table}

\noindent {\bf Results on open world object detection.}
We also compare OpenDet with ORE in the task1 of open world object detection.
As shown in Tab.~\ref{tab:owod_t1}, without accessing open-set data in the training set or validation set,
OpenDet outperforms FR-CNN and ORE by a large margin and achieves comparable results with the Oracle.

\begin{table}[t]
    \centering
\small
\resizebox{\linewidth}{!}{%
\begin{tabular}{c|cc|cccc}
\hline
\multirow{2}{*}{Method} & \multicolumn{2}{c|}{use unknown's annotation} & \multirow{2}{*}{WI$_\downarrow$}   & \multirow{2}{*}{AOSE$_\downarrow$}  & \multirow{2}{*}{mAP$_\mathcal{K \uparrow}$}  \\
 & in train set  & in val set &   &    & \\
\hline
\makecell[c]{FR-CNN \\ (Oracle)}   & $\checkmark$  & $\times$     & 4.27  & 6862 & 60.43   \\
\hline
FR-CNN   & $\times$      & $\times$     & 6.03  & 8468 & 58.81   \\
ORE      & $\times$      & $\checkmark$ & 5.11  & 6833 & 58.93   \\
{\bf OpenDet}  & $\times$      & $\times$     & {\bf 4.44} & {\bf 5781} & {\bf 59.01}   \\
\hline
\end{tabular}%
}
    \vspace{-1mm}
    \caption{{\bf Results on open world object detection~\cite{joseph2021towards}.}}
    \label{tab:owod_t1}
    \end{table}

\section{Comparison with DS~\cite{miller2018dropout}}
\noindent {\bf Implementation details.} DS averages multiple runs of a dropout-enabled model to produce more confident preditions.
As DS has no public implementation, we implement it based on the FR-CNN~\cite{ren2017faster} framework.
Specifically, we insert a dropout layer to the second-last layer of the classification branch in R-CNN, and set the dropout probability to 0.5.
Previous works~\cite{dhamija2020overlooked,joseph2021towards} indicate that DS works even worse than the baseline method;
we show it is effective as long as we remove the dropout layer during training, \emph{i.e.}, we only use the dropout layer in the testing phase.
Besides, original DS can only tell what is known, but do not have a metric for the unknown (\emph{e.g.}, the unknown probability in OpenDet).
We give DS the ability to identify unknown by entropy thresholding~\cite{hendrycks2016baseline}.
In detail, we define proposals with the entropy larger than a threshold (\emph{i.e.}, 0.25) as unknown.

\noindent {\bf DS with different \#runs.} DS requires multiple runs for a given image. We report DS with different number of \#runs in Tab.~\ref{tab:dropout_sampling}.
By increasing \#runs, DS shows substantial improvements on AOSE and \mAPK, while the performance on WI becomes worse.
We report DS with 30 \#runs in our main paper, which is consistent with its original paper~\cite{miller2018dropout}.
\vspace{-3mm}
\begin{table}[t]
    \centering
\small
\setlength{\tabcolsep}{2.5mm}{
\begin{tabular}{c|c|cccc}
\hline
Method & \#runs  & WI$_\downarrow$   & AOSE$_\downarrow$  & mAP$_\mathcal{K \uparrow}$   & AP$_\mathcal{U \uparrow}$  \\
\hline
FR-CNN  & 1 & 18.39 & 15118 & 58.45 & 0     \\
\hline
\multirow{5}{*}{DS} &
1        & 15.26 & 18227 & 56.60 & 5.67  \\
&3       & 16.41 & 14593 & 57.88 & 5.48  \\
&5       & 16.76 & 13862 & 57.98 & 5.31  \\
&10      & 16.91 & 13327 & 58.24 & 4.97  \\
& \cg 30 & \cg 16.98 & \cg 12868 & \cg 58.35 & \cg 5.13  \\
&50      & 17.01 & 12757 & 58.29 & 4.94  \\
\hline
OpenDet& 1 & {\bf 14.95} & {\bf 11286} & {\bf 58.75} & {\bf 14.93} \\
\hline
\end{tabular}%
}
    \caption{{\bf Comparison with DS~\cite{miller2018dropout}.} 
    \#runs denotes the number of runs used for ensemble.
    The row with \textcolor{gray}{gray} background is reported in our main paper.
    }
    \vspace{-3mm}
    \label{tab:dropout_sampling}
    \end{table}

\section{More Qualitative Results.}
\begin{figure}[h]
    \includegraphics[width=\linewidth]{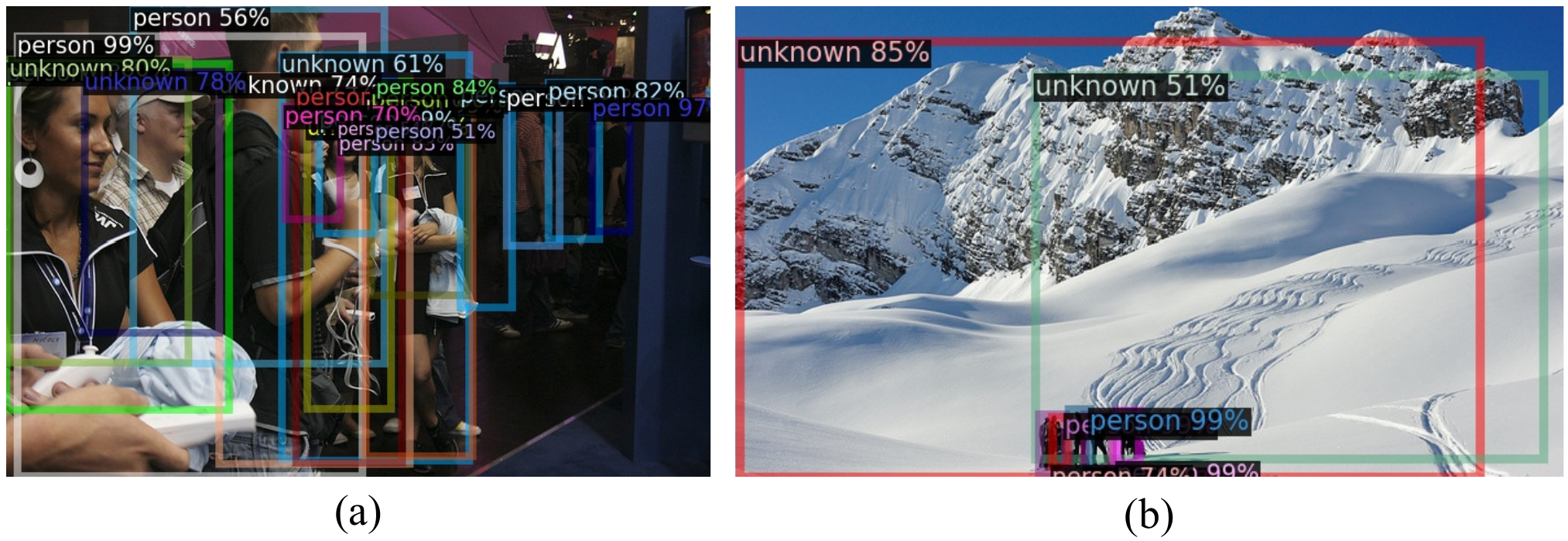}
    \vspace{-5mm}
    \caption{{\bf Failure cases.}}
    \label{fig:failure_cases}
\end{figure}

\begin{figure*}[h!]
    \centering
    \includegraphics[width=0.95\textwidth]{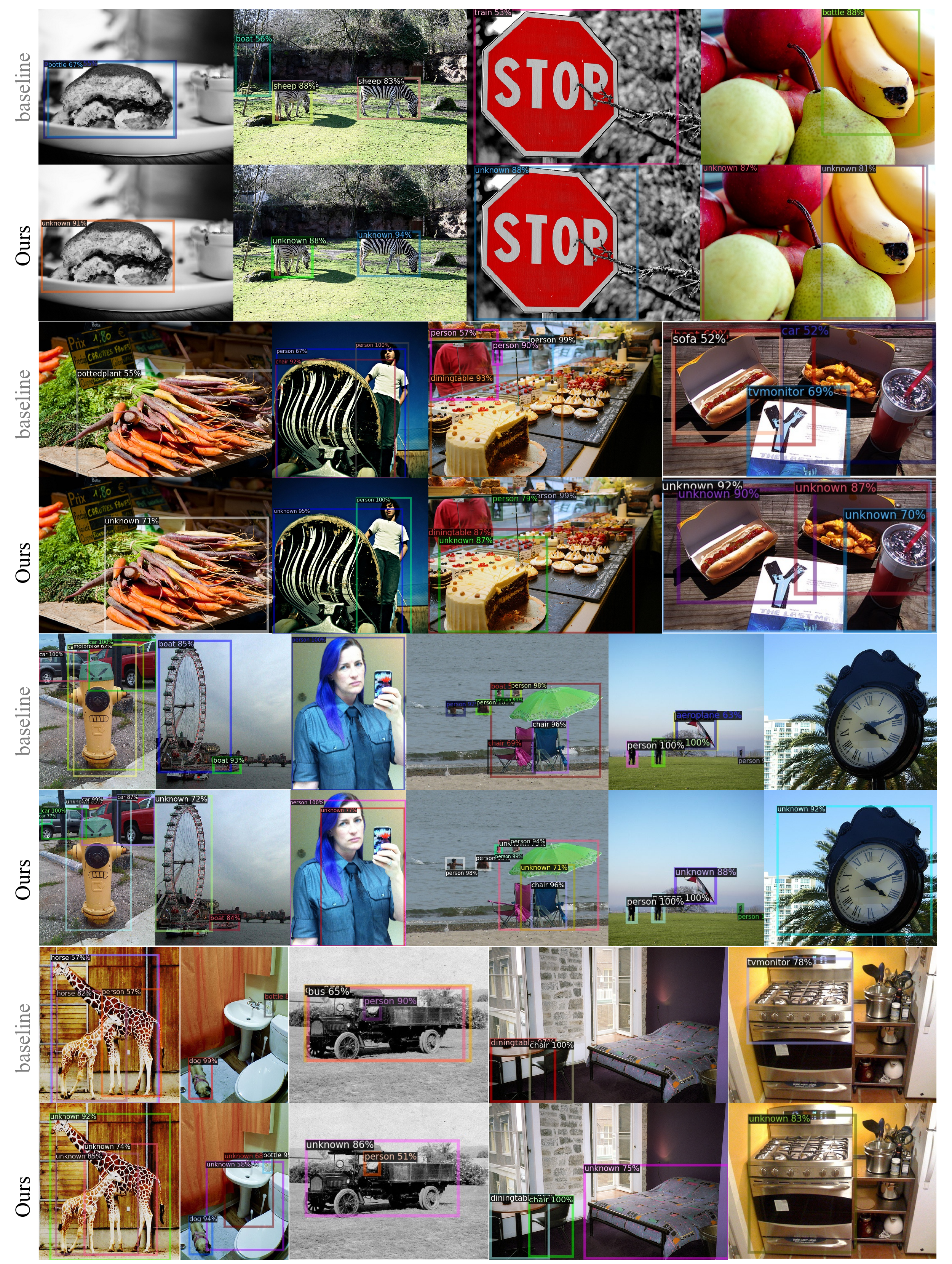}
    \caption{{\bf More qualitative comparisons} between the baseline and OpenDet.}
    \label{fig:more_qualitative_results}
\end{figure*}

Fig.~\ref{fig:more_qualitative_results} gives more qualitative comparisons between the baseline method and OpenDet.
OpenDet can recall unknown objects from known classes and the "background".
Besides, we also give two failure cases in Fig.~\ref{fig:failure_cases}.
{\bf (a)} We find OpenDet performs poorly in some scenes with dense objects, \emph{e.g.}, images with lots of \texttt{person}.
{\bf (b)} OpenDet classifies "real" background to the \texttt{unknown} class.

\newpage
{\small
\bibliographystyle{ieee_fullname}
\bibliography{egbib}

\newcommand{\noop}[1]{}
\begin{thebibliography}{10}\itemsep=-1pt

\bibitem{bendale2015towards}
Abhijit Bendale and Terrance Boult.
\newblock Towards open world recognition.
\newblock In {\em CVPR}, pages 1893--1902, 2015.

\bibitem{bendale2016towards}
Abhijit Bendale and Terrance~E Boult.
\newblock Towards open set deep networks.
\newblock In {\em CVPR}, pages 1563--1572, 2016.

\bibitem{carion2020end}
Nicolas Carion, Francisco Massa, Gabriel Synnaeve, Nicolas Usunier, Alexander
  Kirillov, and Sergey Zagoruyko.
\newblock End-to-end object detection with transformers.
\newblock In {\em ECCV}, pages 213--229. Springer, 2020.

\bibitem{caron2020unsupervised}
Mathilde Caron, Ishan Misra, Julien Mairal, Priya Goyal, Piotr Bojanowski, and
  Armand Joulin.
\newblock Unsupervised learning of visual features by contrasting cluster
  assignments.
\newblock {\em arXiv preprint arXiv:2006.09882}, 2020.

\bibitem{chapelle2006semi}
Olivier Chapelle, Bernhard Scholkopf, and Alexander Zien.
\newblock Semi-supervised learning. 2006.
\newblock {\em Cambridge, Massachusettes: The MIT Press View Article}, 2006.

\bibitem{chen2021adversarial}
Guangyao Chen, Peixi Peng, Xiangqian Wang, and Yonghong Tian.
\newblock Adversarial reciprocal points learning for open set recognition.
\newblock {\em arXiv preprint arXiv:2103.00953}, 2021.

\bibitem{chen2020learning}
Guangyao Chen, Limeng Qiao, Yemin Shi, Peixi Peng, Jia Li, Tiejun Huang,
  Shiliang Pu, and Yonghong Tian.
\newblock Learning open set network with discriminative reciprocal points.
\newblock In {\em ECCV}, pages 507--522. Springer, 2020.

\bibitem{chen2020simple}
Ting Chen, Simon Kornblith, Mohammad Norouzi, and Geoffrey Hinton.
\newblock A simple framework for contrastive learning of visual
  representations.
\newblock In {\em ICML}, pages 1597--1607, 2020.

\bibitem{chen2021exploring}
Xinlei Chen and Kaiming He.
\newblock Exploring simple siamese representation learning.
\newblock In {\em CVPR}, pages 15750--15758, 2021.

\bibitem{cui2021parametric}
Jiequan Cui, Zhisheng Zhong, Shu Liu, Bei Yu, and Jiaya Jia.
\newblock Parametric contrastive learning.
\newblock {\em arXiv preprint arXiv:2107.12028}, 2021.

\bibitem{devries2018learning}
Terrance DeVries and Graham~W Taylor.
\newblock Learning confidence for out-of-distribution detection in neural
  networks.
\newblock {\em arXiv preprint arXiv:1802.04865}, 2018.

\bibitem{dhamija2020overlooked}
Akshay Dhamija, Manuel Gunther, Jonathan Ventura, and Terrance Boult.
\newblock The overlooked elephant of object detection: Open set.
\newblock In {\em WACV}, pages 1021--1030, 2020.

\bibitem{ding2021unsupervised}
Jian Ding, Enze Xie, Hang Xu, Chenhan Jiang, Zhenguo Li, Ping Luo, and Gui-Song
  Xia.
\newblock Unsupervised pretraining for object detection by patch
  reidentification.
\newblock {\em arXiv preprint arXiv:2103.04814}, 2021.

\bibitem{everingham2010pascal}
Mark Everingham, Luc Van~Gool, Christopher~KI Williams, John Winn, and Andrew
  Zisserman.
\newblock The pascal visual object classes (voc) challenge.
\newblock {\em IJCV}, 88(2):303--338, 2010.

\bibitem{gal2016dropout}
Yarin Gal and Zoubin Ghahramani.
\newblock Dropout as a bayesian approximation: Representing model uncertainty
  in deep learning.
\newblock In {\em ICML}, pages 1050--1059. PMLR, 2016.

\bibitem{ge2017generative}
ZongYuan Ge, Sergey Demyanov, Zetao Chen, and Rahil Garnavi.
\newblock Generative openmax for multi-class open set classification.
\newblock In {\em BMVC}, 2017.

\bibitem{girshick2014rich}
Ross Girshick, Jeff Donahue, Trevor Darrell, and Jitendra Malik.
\newblock Rich feature hierarchies for accurate object detection and semantic
  segmentation.
\newblock In {\em CVPR}, pages 580--587, 2014.

\bibitem{grandvalet2004semi}
Yves Grandvalet and Yoshua Bengio.
\newblock Semi-supervised learning by entropy minimization.
\newblock {\em NeurIPS}, 17, 2004.

\bibitem{grill2020bootstrap}
Jean-Bastien Grill, Florian Strub, Florent Altch{\'e}, Corentin Tallec,
  Pierre~H Richemond, Elena Buchatskaya, Carl Doersch, Bernardo~Avila Pires,
  Zhaohan~Daniel Guo, Mohammad~Gheshlaghi Azar, et~al.
\newblock Bootstrap your own latent: A new approach to self-supervised
  learning.
\newblock {\em arXiv preprint arXiv:2006.07733}, 2020.

\bibitem{he2020momentum}
Kaiming He, Haoqi Fan, Yuxin Wu, Saining Xie, and Ross Girshick.
\newblock Momentum contrast for unsupervised visual representation learning.
\newblock In {\em CVPR}, pages 9729--9738, 2020.

\bibitem{he2016resnet}
Kaiming He, Xiangyu Zhang, Shaoqing Ren, and Jian Sun.
\newblock Deep residual learning for image recognition.
\newblock In {\em CVPR}, pages 770--778, 2016.

\bibitem{hendrycks2016baseline}
Dan Hendrycks and Kevin Gimpel.
\newblock A baseline for detecting misclassified and out-of-distribution
  examples in neural networks.
\newblock {\em arXiv preprint arXiv:1610.02136}, 2016.

\bibitem{hwang2021exemplar}
Jaedong Hwang, Seoung~Wug Oh, Joon-Young Lee, and Bohyung Han.
\newblock Exemplar-based open-set panoptic segmentation network.
\newblock In {\em CVPR}, pages 1175--1184, 2021.

\bibitem{jain2014multi}
Lalit~P Jain, Walter~J Scheirer, and Terrance~E Boult.
\newblock Multi-class open set recognition using probability of inclusion.
\newblock In {\em ECCV}, pages 393--409. Springer, 2014.

\bibitem{joseph2021towards}
KJ Joseph, Salman Khan, Fahad~Shahbaz Khan, and Vineeth~N Balasubramanian.
\newblock Towards open world object detection.
\newblock In {\em CVPR}, pages 5830--5840, 2021.

\bibitem{junior2017nearest}
Pedro R~Mendes J{\'u}nior, Roberto~M De~Souza, Rafael de~O Werneck, Bernardo~V
  Stein, Daniel~V Pazinato, Waldir~R de Almeida, Ot{\'a}vio~AB Penatti, Ricardo
  da~S Torres, and Anderson Rocha.
\newblock Nearest neighbors distance ratio open-set classifier.
\newblock {\em Machine Learning}, 106(3):359--386, 2017.

\bibitem{khosla2020supervised}
Prannay Khosla, Piotr Teterwak, Chen Wang, Aaron Sarna, Yonglong Tian, Phillip
  Isola, Aaron Maschinot, Ce Liu, and Dilip Krishnan.
\newblock Supervised contrastive learning.
\newblock {\em arXiv preprint arXiv:2004.11362}, 2020.

\bibitem{kong2021opengan}
Shu Kong and Deva Ramanan.
\newblock Opengan: Open-set recognition via open data generation.
\newblock {\em arXiv preprint arXiv:2104.02939}, 2021.

\bibitem{lakshminarayanan2016simple}
Balaji Lakshminarayanan, Alexander Pritzel, and Charles Blundell.
\newblock Simple and scalable predictive uncertainty estimation using deep
  ensembles.
\newblock {\em arXiv preprint arXiv:1612.01474}, 2016.

\bibitem{lin2017feature}
Tsung-Yi Lin, Piotr Doll{\'a}r, Ross Girshick, Kaiming He, Bharath Hariharan,
  and Serge Belongie.
\newblock Feature pyramid networks for object detection.
\newblock In {\em CVPR}, pages 2117--2125, 2017.

\bibitem{lin2017focal}
Tsung-Yi Lin, Priya Goyal, Ross Girshick, Kaiming He, and Piotr Doll{\'a}r.
\newblock Focal loss for dense object detection.
\newblock In {\em ICCV}, pages 2980--2988, 2017.

\bibitem{lin2014microsoft}
Tsung-Yi Lin, Michael Maire, Serge Belongie, James Hays, Pietro Perona, Deva
  Ramanan, Piotr Doll{\'a}r, and C~Lawrence Zitnick.
\newblock Microsoft coco: Common objects in context.
\newblock In {\em ECCV}, pages 740--755. Springer, 2014.

\bibitem{liu2021swin}
Ze Liu, Yutong Lin, Yue Cao, Han Hu, Yixuan Wei, Zheng Zhang, Stephen Lin, and
  Baining Guo.
\newblock Swin transformer: Hierarchical vision transformer using shifted
  windows.
\newblock {\em arXiv preprint arXiv:2103.14030}, 2021.

\bibitem{malinin2018predictive}
Andrey Malinin and Mark Gales.
\newblock Predictive uncertainty estimation via prior networks.
\newblock {\em arXiv preprint arXiv:1802.10501}, 2018.

\bibitem{miller2019evaluating}
Dimity Miller, Feras Dayoub, Michael Milford, and Niko S{\"u}nderhauf.
\newblock Evaluating merging strategies for sampling-based uncertainty
  techniques in object detection.
\newblock In {\em ICRA}, pages 2348--2354. IEEE, 2019.

\bibitem{miller2018dropout}
Dimity Miller, Lachlan Nicholson, Feras Dayoub, and Niko S{\"u}nderhauf.
\newblock Dropout sampling for robust object detection in open-set conditions.
\newblock In {\em ICRA}, pages 3243--3249. IEEE, 2018.

\bibitem{neal2018open}
Lawrence Neal, Matthew Olson, Xiaoli Fern, Weng-Keen Wong, and Fuxin Li.
\newblock Open set learning with counterfactual images.
\newblock In {\em ECCV}, pages 613--628, 2018.

\bibitem{oza2019c2ae}
Poojan Oza and Vishal~M Patel.
\newblock C2ae: Class conditioned auto-encoder for open-set recognition.
\newblock In {\em CVPR}, pages 2307--2316, 2019.

\bibitem{padhy2020revisiting}
Shreyas Padhy, Zachary Nado, Jie Ren, Jeremiah Liu, Jasper Snoek, and Balaji
  Lakshminarayanan.
\newblock Revisiting one-vs-all classifiers for predictive uncertainty and
  out-of-distribution detection in neural networks.
\newblock {\em arXiv preprint arXiv:2007.05134}, 2020.

\bibitem{redmon2016you}
Joseph Redmon, Santosh Divvala, Ross Girshick, and Ali Farhadi.
\newblock You only look once: Unified, real-time object detection.
\newblock In {\em CVPR}, pages 779--788, 2016.

\bibitem{ren2018meta}
Mengye Ren, Eleni Triantafillou, Sachin Ravi, Jake Snell, Kevin Swersky,
  Joshua~B Tenenbaum, Hugo Larochelle, and Richard~S Zemel.
\newblock Meta-learning for semi-supervised few-shot classification.
\newblock {\em ICLR}, 2018.

\bibitem{ren2017faster}
Shaoqing Ren, Kaiming He, Ross Girshick, and Jian Sun.
\newblock {Faster R-CNN}: Towards real-time object detection with region
  proposal networks.
\newblock {\em IEEE TPAMI}, pages 1137--1149, 2017.

\bibitem{scheirer2012toward}
Walter~J Scheirer, Anderson de Rezende~Rocha, Archana Sapkota, and Terrance~E
  Boult.
\newblock Toward open set recognition.
\newblock {\em IEEE TPAMI}, 35(7):1757--1772, 2012.

\bibitem{scheirer2014probability}
Walter~J Scheirer, Lalit~P Jain, and Terrance~E Boult.
\newblock Probability models for open set recognition.
\newblock {\em IEEE TPAMI}, 36(11):2317--2324, 2014.

\bibitem{sensoy2018evidential}
Murat Sensoy, Lance Kaplan, and Melih Kandemir.
\newblock Evidential deep learning to quantify classification uncertainty.
\newblock {\em arXiv preprint arXiv:1806.01768}, 2018.

\bibitem{sun2021fsce}
Bo Sun, Banghuai Li, Shengcai Cai, Ye Yuan, and Chi Zhang.
\newblock Fsce: Few-shot object detection via contrastive proposal encoding.
\newblock In {\em CVPR}, pages 7352--7362, 2021.

\bibitem{sun2020conditional}
Xin Sun, Zhenning Yang, Chi Zhang, Keck-Voon Ling, and Guohao Peng.
\newblock Conditional gaussian distribution learning for open set recognition.
\newblock In {\em CVPR}, pages 13480--13489, 2020.

\bibitem{tian2019fcos}
Zhi Tian, Chunhua Shen, Hao Chen, and Tong He.
\newblock Fcos: Fully convolutional one-stage object detection.
\newblock In {\em ICCV}, pages 9627--9636, 2019.

\bibitem{van2021unsupervised}
Wouter Van~Gansbeke, Simon Vandenhende, Stamatios Georgoulis, and Luc Van~Gool.
\newblock Unsupervised semantic segmentation by contrasting object mask
  proposals.
\newblock {\em arXiv preprint arXiv:2102.06191}, 2021.

\bibitem{vaze2021open}
Sagar Vaze, Kai Han, Andrea Vedaldi, and Andrew Zisserman.
\newblock Open-set recognition: A good closed-set classifier is all you need.
\newblock {\em arXiv preprint arXiv:2110.06207}, 2021.

\bibitem{wang2021contrastive}
Peng Wang, Kai Han, Xiu-Shen Wei, Lei Zhang, and Lei Wang.
\newblock Contrastive learning based hybrid networks for long-tailed image
  classification.
\newblock In {\em CVPR}, pages 943--952, 2021.

\bibitem{wang2021exploring}
Wenguan Wang, Tianfei Zhou, Fisher Yu, Jifeng Dai, Ender Konukoglu, and Luc
  Van~Gool.
\newblock Exploring cross-image pixel contrast for semantic segmentation.
\newblock {\em arXiv preprint arXiv:2101.11939}, 2021.

\bibitem{wang2020frustratingly}
Xin Wang, Thomas~E Huang, Trevor Darrell, Joseph~E Gonzalez, and Fisher Yu.
\newblock Frustratingly simple few-shot object detection.
\newblock {\em arXiv preprint arXiv:2003.06957}, 2020.

\bibitem{wu2019detectron2}
Yuxin Wu, Alexander Kirillov, Francisco Massa, Wan-Yen Lo, and Ross Girshick.
\newblock Detectron2.
\newblock \url{https://github.com/facebookresearch/detectron2}, 2019.

\bibitem{yoshihashi2019classification}
Ryota Yoshihashi, Wen Shao, Rei Kawakami, Shaodi You, Makoto Iida, and Takeshi
  Naemura.
\newblock Classification-reconstruction learning for open-set recognition.
\newblock In {\em CVPR}, pages 4016--4025, 2019.

\bibitem{zhang2016sparse}
He Zhang and Vishal~M Patel.
\newblock Sparse representation-based open set recognition.
\newblock {\em IEEE TPAMI}, 39(8):1690--1696, 2016.

\bibitem{zhou2021learning}
Da-Wei Zhou, Han-Jia Ye, and De-Chuan Zhan.
\newblock Learning placeholders for open-set recognition.
\newblock In {\em CVPR}, pages 4401--4410, 2021.

\end{thebibliography}
}

\end{document}